%% file: main.tex
\documentclass[conference]{IEEEtran}
\usepackage{times}

\let\labelindent\relax
\usepackage{enumitem}
\usepackage[numbers]{natbib}
\usepackage{multicol}
\usepackage[bookmarks=true]
{hyperref}

\usepackage{bm}
\usepackage{amsmath}
\usepackage{amssymb}
\usepackage{mathtools} % Provides \coloneqq
\usepackage{amsfonts}
\usepackage{comment}
\usepackage{graphicx}
\usepackage[dvipsnames]{xcolor}

\usepackage{gensymb}

\usepackage[linesnumbered,ruled,vlined]{algorithm2e}
\newcommand{\code}[1]{\textcolor{RoyalBlue}{$\rhd$ \emph{#1}}}
\makeatletter
\renewcommand{\algocf@Vline}[1]{%
\strut\par\nointerlineskip
\algocf@push{\skiprule}%
\hbox{\vrule
  \vtop{\algocf@push{\skiptext}%
    \vtop{\algocf@addskiptotal #1}\Hlne}}%
\algocf@pop{\skiprule}%
\nointerlineskip %
}%
\makeatother

\usepackage{booktabs}
\usepackage{makecell}  % for \Xhline
\usepackage{tabularx} % For controlling table width
\usepackage{cases}  % provides the numcases environment

\usepackage{pifont} % for cross symbol

\usepackage{multirow}

\usepackage{caption}
\usepackage{svg}  % for svg type image

\usepackage{amssymb}
\usepackage{booktabs} % for pandas-generated table
\usepackage{dblfloatfix}

\usepackage{todonotes} % add todos

\begin{document}

% paper title

% \title{Contrastive Diffusion Policy Learning }
% \title{Contrastive Diffusion policy: Learning Action Diffusion through Corrections }
% \title{Contrastive Diffusion Policy: Learning Action-Chunking Diffusion through Corrections }

\title{Set-Supervised Diffusion Policy: Learning Action-Chunking Diffusion through Corrections }
% \title{Learning Diffusion Policies from Failures and Corrections}
% \title{Contrastive Diffusion policy: Learning Diffusion Policies from Contrastive Actions}

% \title{Diffusion policy from human corrections: Learning Action Diffusion through Interactive Imitation }

% You will get a Paper-ID when submitting a pdf file to the conference system
% \author{Author Names Omitted for Anonymous Review. Paper-ID 382}
\author{
    \IEEEauthorblockN{
        Zhaoting Li$^{1}$, Gang Chen$^{1}$, Javier Alonso-Mora$^{1}$, 
        Cosimo Della Santina$^{1}$, Jens Kober$^{2}$
    }
    % \IEEEauthorblockA{
    %     $^{1}$Delft University of Technology,
    %     $^{2}$University of Stuttgart
    % }
    \IEEEauthorblockA{
        $^{1}$Delft University of Technology,
        $^{2}$University of Stuttgart\\
        \{Z.Li-23, G.Chen-5, J.AlonsoMora, C.DellaSantina\}@tudelft.nl,
        jens.kober@ki.uni-stuttgart.de
    }
}

\IEEEpeerreviewmaketitle
% \maketitle
\twocolumn[{%
	\renewcommand\twocolumn[1][]{#1}%
	\maketitle
        \vspace{-6mm}
	\begin{center}
		\includegraphics[width=0.99\textwidth]{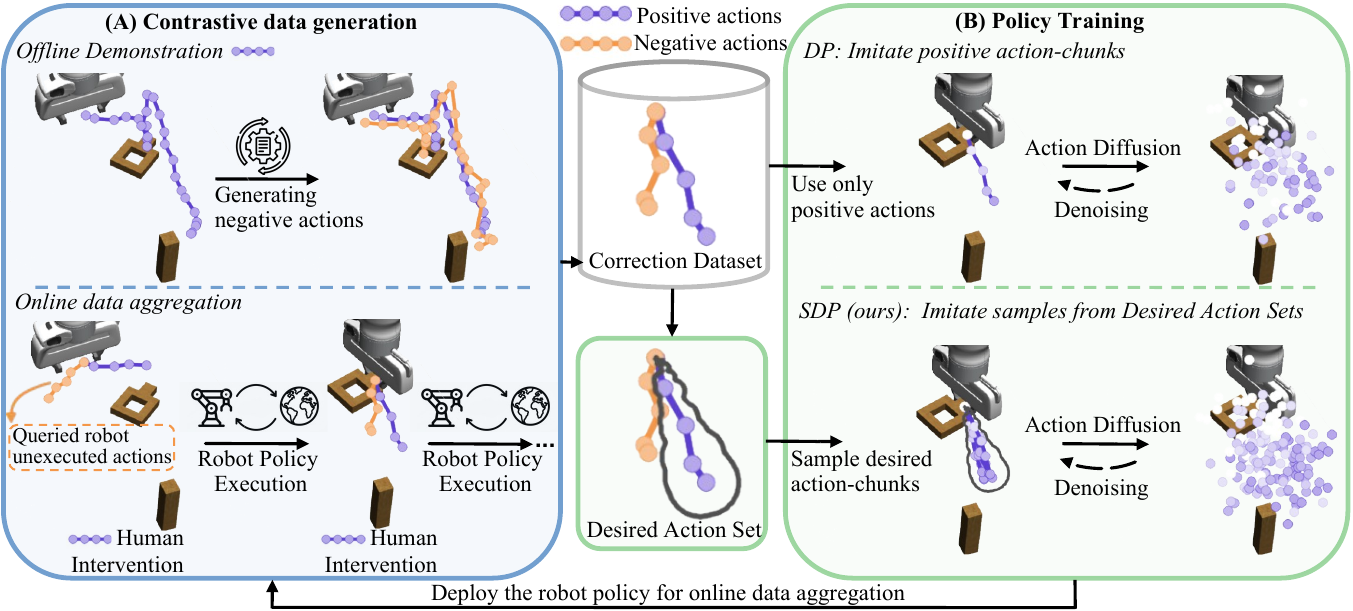}
% \includesvg[width=0.99\textwidth,inkscapelatex=false]{figs/Fig_3_cover_Figure_3.svg}
 \captionof{figure}{\small
 Framework of our Set-Supervised Diffusion Policy (SDP) method. A: Contrastive data, consisting of positive-negative action-chunk pairs, can be generated for both offline demonstrations and online interventions.
 B: SDP uses this contrastive data to construct desired action sets, which serve as set-valued action targets for policy learning.
 The diffusion policy is trained to generate action-chunks within these sets. 
\label{fig:framework}}
	\end{center}
}]
% \begin{abstract} Behavior cloning (BC) traditionally relies on demonstration feedback, assuming the demonstrated actions are optimal. This assumption can lead to overfitting, particularly with expressive models like the energy-based model utilized in Implicit BC. To address this, we reformulate behavior cloning as an optimal action estimation problem and introduce \textit{Contrastive policy Learning from Interactive Corrections (CLIC)}. CLIC leverages human corrections—both absolute and relative—to construct a set of desired actions and optimizes the policy to select actions from this set. We provide theoretical guarantees for the convergence of the desired action set to optimal actions in both single and multiple optimal action cases. Extensive simulation and real-robot experiments validate CLIC's advantages over state-of-the-art BC methods, including stable training of energy-based models, robustness to feedback noise, and adaptability to diverse feedback beyond demonstrations.
% To our best knowledge, this work offers a fresh perspective on policy learning by focusing on iteratively estimating optimal actions rather than directly imitating them. 
% The code will be publicly available. 
% \end{abstract}
\begin{abstract} 
Diffusion policies have recently emerged as a powerful framework for robotic manipulation. However, like other behavior cloning methods, they remain vulnerable to distributional shift, often requiring human-in-the-loop interventions to correct failures during deployment. These interactions naturally provide paired supervision in the form of the robot’s undesired actions and the human teacher’s corrective actions. Yet existing data aggregation pipelines and standard behavior cloning losses largely ignore this negative signal from undesired actions, leading to overfitting to teacher's actions and an increasing reliance on costly expert data. To address this limitation, we propose Set-Supervised Diffusion Policy (SDP), a novel learning framework that utilizes contrastive action-chunk data to train diffusion policies from human corrections. From paired positive and negative action-chunks, SDP constructs a set of desired action-chunks and designs a training pipeline that encourages the diffusion policy to align with the set. Through extensive experiments across multiple robotic manipulation tasks, we demonstrate that SDP consistently improves policy performance, with particularly strong gains in robustness to noisy data. Moreover, SDP induces high-quality aggregated datasets, enabling more efficient and reliable policy learning from human-in-the-loop corrections. Our code is available at \href{https://set-supervised-diffusion-policy.github.io/ }{https://set-supervised-diffusion-policy.github.io/}.
\end{abstract}

% \begin{keywords}
%     Interactive Imitation Learning, Corrective feedback, Contrastive Learning, Learning from Demonstration, Energy-based Models
% \end{keywords}

% \begin{IEEEkeywords}
% Interactive Imitation Learning, Learning from Demonstration, Diffusion Policies, Data Aggregation
% \end{IEEEkeywords}

\input{contents/01_introduction}

\input{contents/02_Related_work}

\input{contents/03_Preliminaries}

\input{contents/04_Methodology}

\input{contents/05_Experiment}

% \input{Discussion}
\input{contents/06_Conclusion}

\section*{Acknowledgments}
This project is made possible by a contribution from the National Growth Fund program NXTGEN Hightech.
We also acknowledge the Delft AI Cluster (DAIC) for providing computational resources. We'd like to thank Rodrigo P{\'e}rez-Dattari for helpful discussions, and Qifan Luo and Zi Huang for their valuable feedback on the code repository. 
%% Use plainnat to work nicely with natbib. 

% % \bibliographystyle{plainnat}
% \bibliographystyle{unsrt} % Ensures citations appear in order
\bibliographystyle{plainnat}
\bibliography{main}

% \bibliographystyle{./IEEEtranBST/IEEEtran}
% \bibliography{./IEEEtranBST/IEEEabrv, references}

% \input{rebuttal}

\newpage
$\quad$
% TO remove the newpages
\newpage
\input{contents/07_Appendix}

% \newpage
% $\quad$

% \newpage
% \input{contents/To_remove_Rebuttal}

\end{document}

%% file: contents/01_introduction.tex
\section{Introduction}
\label{sec:introduction}

% Start of intro: Diffusion Policy (DP) is good, and data collected during deployment (intervention) can help it reduce covariate shift.  

% Key problem: however, these intervention data are not fully utilized: (1) most only record intervention action and do not save the corresponding robot action; (2) even record robot action, the BC algorithm only imitates the teacher action, causing the waste of robot action data. This also makes BC algorithm rely on expert demonstrations.

% Desire: The algorithm should utilize both intervention action and robot action effectively.

% Obstacle: (1) DPO (or other preference learning algorithms) can be applied to encourage teacher action and discourage the robot action. However, DPO is inefficient in a continuous action space and relies on a base policy with good performance. 
% (2) CLIC can learn from corrections, but it utilizes EBM, which is hard to train with action chunks.

% Our idea: In this work, we address these challenges by proposing Contrastive Diffusion Policy -- an algorithm that incorporates the power of CLIC into diffusion policies

Imitation Learning (IL) has established itself as a cornerstone paradigm for robotic manipulation, enabling agents to acquire complex skills directly from expert demonstrations. 
Within this landscape, Diffusion Policies (DP) have emerged as a powerful policy representation, leveraging the generative diffusion models to operate robustly in high-dimensional continuous action-chunk spaces \cite{2023_diffusionpolicy, 2023_score_diffusion_policy, 2025RSS_pi0_flow_model}. 
Despite these strengths, DPs remain vulnerable to distributional shift inherent in Behavior Cloning (BC), where the state distribution induced by the policy drifts away from the training data \cite{2024_Diffusion_dagger, 2011_DAgger}. A widely adopted mitigation strategy is human-in-the-loop intervention, where a human teacher temporarily takes control to correct robot's mistakes \cite{2019_HG_DAgger, 2020_RSS_expert_interventio_learning}.  However, current BC approaches fail to fully exploit this correction information.

While these corrections generate a rich stream of data containing both the robot’s undesired actions and the teacher’s intervention actions, standard BC objectives only maximize the likelihood of the teacher’s actions, but ignore information from the robot’s undesired actions. 
Consequently, existing data aggregation methods record only the teacher's actions \cite{2019_HG_DAgger, 2011_DAgger, 2024_Diffusion_dagger},
biasing learning toward the teacher's distribution 
\cite{2025_IL_Recovery_Correction}.
This bias can cause policies to overfit and over-rely on costly high-quality interventions \cite{2025_CLIC}.

Prior work has sought to address this limitation through methods that learn from preferences or corrections.
However, approaches such as Direct Preference Optimization (DPO) struggle with high-dimensional action-chunk spaces and require a strong pre-trained policy \cite{chen_deformpam_2025, 2023_DPO}.
Beyond preference-based learning, CLIC~\cite{2025_CLIC} proposes set-valued supervision for corrective feedback, where a positive-negative correction pair defines a set of acceptable actions rather than a single action target.
This relaxation helps mitigate BC's tendency to overfit to imperfect corrections.
 Yet, CLIC employs Energy-Based Models (EBMs) as the policy representation \cite{2020_Score_based_diffusion}, which are difficult to train stably and integrate with action-chunking—one of the key mechanisms underpinning the strong empirical performance of DPs \cite{2023_diffusionpolicy}.

% In this work, we introduce Set-Supervised Diffusion Policy (SDP), a new framework that enables diffusion policies to learn from both teacher's actions and the robot's undesired actions. 
% In this work, we introduce Set-Supervised Diffusion Policy (SDP), a framework that trains diffusion policies from correction-consistent action sets constructed using both the teacher's corrective actions and the robot's undesired actions.
In this work, we introduce Set-Supervised Diffusion Policy (SDP), a framework that trains diffusion policies with set-valued action targets constructed from both the teacher's corrective actions and the robot's undesired actions.
To achieve this, we embed the insights from the set-valued supervision of CLIC into diffusion policies and propose a set of desired action-chunks from a single pair of contrastive action-chunk data.
Intuitively, this desired action set provides information about alternative action-chunks that could be equally desirable as the human action. However, directly applying the CLIC loss to train diffusion policies with the desired set is non-trivial, as the loss relies on explicit probability estimation available in EBMs but absent in diffusion models.
To address this, we generalize the CLIC loss by sampling desired action-chunks from the desired set and maximizing the likelihood of these samples.
Extensive experiments show that SDP consistently learns stronger policies in both online and offline settings, while also producing higher-quality training data during online data aggregation. Together, these results demonstrate that contrastive supervision is a powerful yet largely underutilized signal for robot policy learning.

%% file: contents/02_Related_work.tex
\section{Related Work}
\label{sec:related_work}

% \subsection{Diffusion Policies, Flow Matching, and Behavior Cloning}
\noindent\textbf{Diffusion Policies:}
Diffusion Policies (DP) have become a leading approach for visuomotor imitation learning thanks to their ability to generate expressive, multi-modal actions \cite{2023_diffusionpolicy, 2023_score_diffusion_policy}. 
Prior work has extended DP with 3D scene representations \cite{2024_Rss_DP3}, SE(3)-equivariant architectures \cite{2025corl_equibot_DP}, hierarchical formulations \cite{2025_HDP}, and flow-matching objectives to improve inference efficiency \cite{2025corl_ManiFlow, 2025corl_flower}.
Generalist vision-language-action (VLA) policies similarly adopt diffusion or flow-based action heads for their expressivity \cite{2024_Octo, 2025RSS_pi0_flow_model, 2025_smolvla}. 
Across these works, diffusion or flow models are trained using the BC objective.

In contrast, our method departs from the BC paradigm and trains DPs using a \emph{set of desired action-chunks}. This idea builds on the set-valued supervision introduced by CLIC,  which trains policies to imitate a set of desired actions inferred from human corrections \cite{2025_CLIC}. Similarly, counterfactual BC proposes a set of actions that the expert could have intended \cite{2025_counterfactual_BC}. However, these approaches consider single-step actions and cannot be applied directly to action-chunks. We address this gap by extending desired-action-set supervision to the space of action-chunks, and designing novel training pipelines to train DP with set-valued supervision.

% \subsection{Intervention, DAgger, and Data Aggregation}
\noindent\textbf{Human Intervention and Data Aggregation:}
Covariate shift is a central challenge in IL, and data aggregation is a well-established strategy to mitigate it \cite{2022_IIL_survey, 2018_D_COACH, 2011_DAgger}. DAgger addresses this issue by iteratively querying an expert to aggregate demonstrations along the learner’s state distribution \cite{2011_DAgger}. Variants have refined how interventions are triggered and incorporated: HG-DAgger emphasizes human-gated interventions \cite{2019_HG_DAgger, 2023_RSS_Robot_Learning_on_the_job, 2023_IIFL_implicit_interactive_BC}, while other work views interventions as implicit corrective signals \cite{2020_RSS_expert_interventio_learning}. 
Other approaches focus data collection on informative states, such as bottleneck states in long-horizon manipulation tasks \cite{2024iros_juicer_IL_furniture}.
Recent work has further integrated DP with intervention-based data aggregation \cite{2025arxiv_intervention_training_Diffusion, 2024_Diffusion_dagger, 2025_IL_Recovery_Correction, xu2026compliant}.

Unlike DAgger-style approaches that rely on BC losses and ignore negative actions, our approach aggregates correction data as paired positive-negative actions and optimizes the policy with set-valued supervision. 
By relaxing pointwise imitation, this supervision encourages exploration within a desired action set, yielding broader state coverage during online learning.

% Unlike DAgger-style approaches that rely on BC losses and discard the robot's negative actions, SDP aggregates correction data as paired positive-negative action-chunks.
% The resulting desired action sets provide set-valued action targets for policy learning, yielding broader state coverage during online learning.

% \subsection{Finetuning}
\noindent\textbf{Finetuning with newly aggregated data:}
Finetuning is widely used to adapt pretrained policies to new tasks, domains, or user preferences. 
Supervised finetuning with additional demonstrations is common in VLA models \cite{2025RSS_pi0_flow_model, 2025_smolvla, 2024_Octo}, but inherits BC’s sensitivity to label noise and distribution shift.
Finetuning with human preference data has also gained traction \cite{2025_ICRA_DP_Preference_PPO, 2024_NIPS_preference_finetuning_data_coverage, 2025arxiv_Action_preference_optimization}, though binary preferences can be sparser than action-level feedback.
Another line of work uses reinforcement learning (RL) \cite{2025NIPS_reinflow, 2025ICLR_DPPO_diffusion}. Human-in-the-loop RL has been shown to improve dexterous manipulation beyond imitation performance \cite{2025_Scirobotics_HIL_RL}. Residual RL methods finetune only a corrective residual on top of a BC policy \cite{2025icra_IL_RL_refinement_furniture}. Low-quality data can also be utilized within an RL-based pipeline \cite{2025arxiv_pi_star6}. 

Different from these approaches, our SDP loss inherits the benefits of set-valued supervision and mitigates the overfitting issue in supervised finetuning. 
By optimizing the policy over a desired action set, SDP also shares a core advantage of RL fine-tuning: it enables policy improvements beyond the quality of the training data, rather than simply imitating it.

%% file: contents/03_Preliminaries.tex
\section{Preliminaries}
\label{sec:Preliminaries}

\subsection{Diffusion Policy and Action Chunks}

Denoising Diffusion Probabilistic Models (DDPMs) \cite{2015_diffusion, 2020_diffusion, 2020_Score_based_diffusion} generate samples by progressively transforming Gaussian noise into a data distribution through a learned denoising process.
Diffusion Policy (DP) \cite{2023_diffusionpolicy} adapts the DDPM formulation to imitation learning by modeling distributions over \emph{action chunks}—sequences of single-step actions $\mathbf{a} \in \mathcal{A}$ conditioned on observations $\mathbf O$. The policy $
    \pi_{\theta}(\mathbf{A}_t | \mathbf{O}_t)$ generates  an action-chunk of fixed horizon $T$
\begin{equation}
    \mathbf{A}_t = \left[\mathbf{a}_t, \mathbf{a}_{t+1}, \dots, \mathbf{a}_{t + T - 1} \right]\in \mathcal{A}^T,
\end{equation}
allowing the policy to resolve multimodal behavior and reduce the covariate shift issue common in robotics tasks.

During inference, DP generates clean actions $\mathbf{A}_t^{0}$ by following the reverse diffusion process of DDPM, conditioned on the current observation $\mathbf{O}_t$. Starting from a Gaussian prior $\mathbf{A}_t^{K} \sim \mathcal{N}(0, I)$, the action is iteratively denoised according to
\begin{equation}
    \mathbf{A}_t^{k-1} = 
    \alpha_k \left(\mathbf{A}_t^k - \gamma_k \, \epsilon_\theta(\mathbf{O}_t, \mathbf{A}_t^{k}, k)\right) 
    + \sigma_k \mathcal{N}(0, I),
    \label{eq:DP_denoising_process}
\end{equation}
for $k = K, \dots, 0$.
Here, $\epsilon_\theta$ denotes a neural network that predicts the injected noise at step $k$.
% conditioned on both the observation and the noisy action. 
The coefficients $\alpha_k$, $\gamma_k$, and $\sigma_k$ are determined by a predefined noise scheduler.
% that controls the signal-to-noise ratio throughout the reverse diffusion process. 
After $K$ denoising steps, the procedure yields the clean action $\mathbf{A}_t^{0}$.

Training of $\epsilon_\theta$ proceeds by sampling clean action data $\mathbf{A}^0_t:=\mathbf{A}_t$ and the observation $\mathbf{O}_t$ from the dataset, selecting a random denoising step $k$, and corrupting the sample using a forward diffusion process:
\begin{equation}
    \mathbf{A}^k_t = \sqrt{\bar{\alpha}_k} \mathbf{A}_t^0 
        \!+ \!\sqrt{1 - \bar{\alpha}_k}\boldsymbol{\epsilon}^k,
\end{equation}
where $\bar{\alpha}_k$ accumulates the noise schedule and $\boldsymbol{\epsilon}^k \sim \mathcal{N}(0,I)$. The training objective is 
% \begin{equation}
%     L_t^\text{DP} 
%     \!\!= \! \mathbb{E}\!_{\substack{k\sim [1, K] \\ \mathbf{O}_t, \mathbf{A}_t^0, \boldsymbol{\epsilon}^k}}
%     \Bigl[
%     \| \boldsymbol{\epsilon}^k \!\!- \!
%     \epsilon_\theta(
%         \mathbf{O}_t,
%         \mathbf{A}^k_t, 
%         k
%     ) \|^2
%     \Bigr].
% \end{equation}
\begin{equation}
\ell_t^\text{DP} 
    \!\!= \! \mathbb{E}_{{k\sim [1, K], \mathbf{O}_t, \mathbf{A}_t^0, \boldsymbol{\epsilon}^k}}
    \big[
    \| \boldsymbol{\epsilon}^k \!\!- \!
    \epsilon_\theta(
        \mathbf{O}_t,
        \mathbf{A}^k_t, 
        k
    ) \|^2
    \big].
    \label{eq:DP_Loss}
\end{equation}
which encourages accurate reconstruction from the corrupted noise. Minimizing this objective maximizes a variational lower bound on the data log-likelihood.

% \subsection{Contrastive Learning from Interactive Corrections (CLIC)}
\subsection{Set-valued Supervision from Interactive Corrections}
\label{sec:Preliminaries:CLIC}

CLIC~\cite{2025_CLIC} introduced a set-valued supervision framework for learning from interactive corrections.
Given a state $\mathbf{s}$, the correction data consists of a positive action $\mathbf{a}^+$ and a negative action $\mathbf{a}^-$.
Negative actions are generated by the suboptimal robot's policy, while positive actions are actions provided by human teachers as feedback signals to improve the robot's policy. 
Rather than treating $\mathbf{a}^+$ as the only valid target,  this formulation constructs a desired action set that includes actions consistent with the correction.
\subsubsection{Single-Step Desired Action Set}
\label{sec:Preliminaries:CLIC:desiredA}
Given $\mathbf{s}$ and an action pair $(\mathbf a^-, \mathbf a^+)$, CLIC constructs a \textit{single-step desired action set} $\hat{\mathcal{A}}^{\mathrm{ss}}(\mathbf a^-, \mathbf a^+) \subseteq\mathcal{A}$,
which contains actions that are consistent with the correction data.
% which provides a \textit{weaker supervisory target} than BC.
This set provides guidance on how to improve the policy by ruling out actions outside it.
BC can be treated as a special case of CLIC with $\hat{\mathcal{A}}^{\mathrm{ss}}(\mathbf a^-, \mathbf a^+)= \{ \mathbf a^+\}$.
However, BC assumes that $\mathbf a^+$ is a perfect training target, and this assumption can be brittle when corrections are noisy or when multiple nearby action-chunks are acceptable.
In contrast, CLIC allows the desired action set to include imperfect actions, assuming only that a subregion of the set contains good actions.
This weaker supervisory target helps reduce overfitting to individual imperfect action labels. 

For $\mathbf{a}^+$ being intervention data,  also referred to as an absolute correction, $\hat{\mathcal{A}}^{\mathrm{ss}}(\mathbf a^-, \mathbf a^+)$ can be defined as a ball centered at $\mathbf{a}^+$:
\begin{equation}
    \hat{\mathcal{A}}^{\mathrm{ss}}{(\mathbf a^-, \mathbf a^+)} = \{ \mathbf a \in \mathcal{A} |r \cdot \mathbb{D}(\mathbf a^+, \mathbf a^-) \geq \mathbb{D}(\mathbf a, \mathbf a^+) \},  
    \label{eq:desired_circular_space}
\end{equation}
where $\mathbb{D}(\mathbf{a}_1,\mathbf{a}_2)=\|\mathbf{a}_1-\mathbf{a}_2\|$, and $r\in[0,1]$ is a radius-ratio hyperparameter controlling the volume of the ball.
Thus, the negative action $\mathbf a^-$ determines the scale of the desired set through its distance to the positive action $\mathbf a^+$.

\subsubsection{Policy shaping via desired action sets}
The desired action set can be utilized to guide the policy improvement. 
The policy can be trained by increasing the probability of selecting actions within the set $\hat{\mathcal{A}}^{ss}(\mathbf{a}^-, \mathbf{a}^+)$, denoted as 
\begin{equation}
    \max_{\smash{\mathbf{\theta}}} 
    \; \pi_{\mathbf \theta}(\mathbf{a} \in \hat{\mathcal{A}}^{\mathrm{ss}}(\mathbf{a}^-, \mathbf{a}^+) | \mathbf{s})
    \label{eq:CLIC_general_objective}
\end{equation}
 % For a policy $\pi_\theta$  represented by energy-based models (EBMs), Eq.~\eqref{eq:CLIC_general_objective} can be rewritten as 
  % For a policy $\pi_\theta$  represented by EBMs, Eq.~\eqref{eq:CLIC_general_objective} can be rewritten as 
  This objective can be achieved by minimizing the KL loss: 
 \begin{align}
    \ell_{KL}(\mathbf \theta) = \!\! \!\!\underset{(\mathbf{s}, \mathbf{a}^{-}, \mathbf{a}^{+}) \sim p_{\mathcal D}}{\mathbb{E}} \!\!\left[ \mathrm{KL}\left(\pi^{\text{target}}(\mathbf a| \mathbf s ) \big\| \pi_{\mathbf \theta}(\mathbf a | \mathbf s) \right) \right], 
    \label{eq:KL_loss_general}
\end{align}
where $\pi^{\text{target}}(\mathbf a| \mathbf s ) $ is calculated by Bayes' rules to assign a higher probability to actions within $\hat{ \mathcal{A}}^{\mathrm{ss}}{(\mathbf a^-, \mathbf a^+)}$.
% Although CLIC has been shown to effectively handle noisy feedback and feedback beyond demonstration (i.e. relative correction), it fails to generate action-chunk outputs due to the limited encoding capabilities of EBMs. 
% In this work, we extend CLIC to the space of action-chunks and design novel loss functions to train diffusion policies with CLIC. 

% \subsubsection{Aggregating Corrections to refine effective supervision region}
\subsubsection{Refining Supervision Targets via Correction Aggregation}

A single desired action set provides a weaker supervisory target than BC: it may contain suboptimal actions and is not assumed to exactly identify the true optimum.
CLIC mitigates this ambiguity by aggregating corrections collected during interaction.
For each state, different corrections rule out different regions of the action space that are inconsistent with the teacher's feedback.
As corrections accumulate, the remaining region consistent with the feedback is progressively refined, yielding a more informative supervision signal.

% In practice, CLIC does not explicitly maintain this refined region for every state.
% Instead, CLIC stores corrections in a replay buffer and repeatedly shapes the policy toward the corresponding single-step desired action sets using Eq.~\eqref{eq:KL_loss_general}.
% Through this iterative process, the policy implicitly accumulates constraints from past corrections and approximates the refined desired region.

%% file: contents/04_Methodology.tex
% \input{contents/04_1_desired_action_space}

% \input{contents/04_2_policy_shaping}

\section{Method}

In this section, we detail Set-Supervised Diffusion Policy (SDP), which trains a diffusion policy with set-valued supervision from human corrections.
Instead of using BC loss to imitate individual action labels, SDP adapts the single-step desired action set formulation introduced in Sec. \ref{sec:Preliminaries:CLIC:desiredA}.
To achieve this, we first extend this single-step desired action set to action-chunks in Sec. \ref{sec:method:desired_action_set}. 
Building on this desired action set for action-chunks, in Sec. \ref{sec:method:policy_learning_desired_action_set}, we then derive a training objective that encourages the diffusion policy to generate action-chunks within this set. Finally, we detail the practical implementation and present the SDP algorithm in Sec. \ref{sec:method:algorithm}.

\subsection{Problem Formulation}
We consider the problem of policy learning from corrections, where a robot learns to perform complex tasks from human interventions.
We assume access to a correction dataset $\mathcal{D}_{+-}=\{[\mathbf{O},\mathbf{A}^- , \mathbf{A}^+]\}$, where $\mathbf{A}^-$ and $\mathbf{A}^+$ are paired negative and positive action-chunks.
Our goal is to leverage $\mathcal{D}_{+-}$ to define a set of desired action-chunks for each observation and to use this information to improve a parameterized policy $\pi_{\theta}(\mathbf{A}|\mathbf{O})$.
We consider two learning settings:
{(i) Online interactive imitation learning}, where corrective feedback is collected during policy execution and aggregated into $\mathcal{D}_{+-}$;
{(ii) Offline learning}, where the policy is trained from a fixed dataset $\mathcal{D}_{+-}$, obtained either from online human interventions or constructed by augmenting offline demonstration datasets.

\begin{figure*}[t!]
	\centering
    \includegraphics[width=0.98\textwidth]{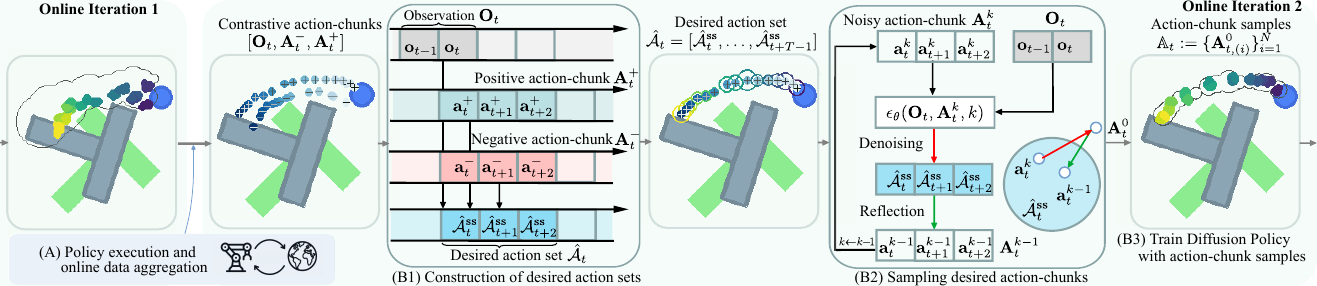}
	% \caption{Overview of CDP. (a) Given contrastive action-chunk pairs, the desired action set is constructed. (b) Using the desired action set, desired action-chunks are sampled using a denoising process with reflections. (c) The diffusion model is trained by imitating these action-chunk samples. (d) The policy interacts with the environment and aggregates new intervention data. The new data leads to a desired action set with a smaller volume. }
    \caption{SDP Overview, illustrated on the Push-T task~\cite{2023_diffusionpolicy}, where the blue end-effector pushes the gray T-object toward the green goal pose.
(A) The policy interacts with the environment to collect new intervention data, progressively shrinking desired action sets. This refined supervision region is also illustrated in Fig. \ref{fig:desired_action_set_visualization} (c).
(B1) Given an observation, one contrastive action-chunk pair defines a desired action set.
(B2) Desired action-chunks are sampled from this set via a denoising process with reflections, producing samples that remain inside the set while staying close to the current policy distribution.
(B3) The diffusion policy is trained to imitate sampled action-chunks.
}
\label{fig:method_overview}
\end{figure*}
% \subsection{Sequence of desired action sets}
\subsection{Extending Desired Action Set to Action-Chunks}
\label{sec:method:desired_action_set}

\begin{figure}[t!]
	\centering
	\includegraphics[width=0.47\textwidth]{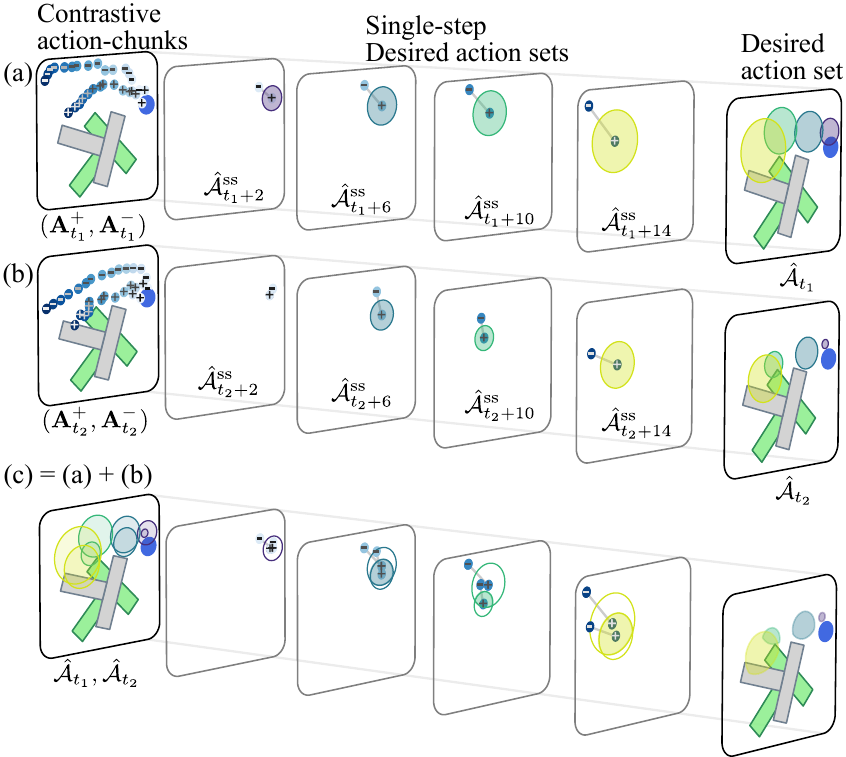}
	\caption{(a)(b): Visualization of the  desired action set. (c): Combination of the desired action sets shown in (a) and (b). New corrections at a state can progressively refine the effective supervision region and reduce the influence of suboptimal actions from earlier desired action sets, such as $\hat{\mathcal{A}}_{t1}$.   }
\label{fig:desired_action_set_visualization}
\end{figure}

CLIC \cite{2025_CLIC} shows that imitating a set of desired actions, constructed by a pair of positive and negative actions, achieves more robust behavior than imitating only the positive action. 
Building on this idea, we generalize the single-step desired action set to action-chunks. 
At step $t$, under observation $\mathbf{O}_t$, we denote a positive and negative action-chunk as
\begin{align*}
    \mathbf{A}^+_t \vcentcolon=[\mathbf a_{t}^+, \mathbf a_{t+1}^+,\dots, \mathbf a_{t + T-1}^+ ], \\
    \mathbf{A}^-_t \vcentcolon=[\mathbf a_{t}^-, \mathbf a_{t+1}^-,\dots, \mathbf a_{t + T-1}^- ].
\end{align*}
Given $\mathbf{O}_t$ and the preceding positive actions $\mathbf a_t^+, \dots, \mathbf a_{t+i-1}^+$, the action $\mathbf a_{t+i}^+$ is better than $\mathbf a_{t+i}^-$ for all $i \in [1, T-1]$.
% DP implicitly assumes that $\mathbf{a}_{t+i}^+$ is optimal given $\mathbf{O}_t$.
% To maintain consistency with DP and enable efficient learning, we adopt the same approximation and assume that
As we consider demonstration data or intervention feedback, to make  learning efficient, we adopt the approximation that $\mathbf a_{t+i}^+$ being a better action than $\mathbf a_{t+i}^-$ depends only on   $\mathbf{O}_t$.
Under this approximation, each timestep within the action-chunk yields a contrastive single-step action pair $(\mathbf a_{t+i}^-, \mathbf a_{t+i}^+)$. Each such pair defines a {single-step desired action set} in CLIC: 
\begin{equation}
\hat{\mathcal A}^{\mathrm{ss}}_{t+i}
\vcentcolon=
\hat{\mathcal A}^{\mathrm{ss}}(\mathbf a_{t+i}^-, \mathbf a_{t+i}^+)
\subseteq \mathcal A.
\end{equation}
We then define the \emph{desired action set} associated with the contrastive action-chunk pair $(\mathbf A_t^+, \mathbf A_t^-)$ as the sequence of single-step desired action sets
\begin{align}
\hat{\mathcal A}_t
\vcentcolon=[\hat{\mathcal A}^{\mathrm{ss}}_t, \hat{\mathcal A}^{\mathrm{ss}}_{t+1}, \dots, \hat{\mathcal A}^{\mathrm{ss}}_{t+T-1}].
\label{eq:desired_action_set_Chunk}
\end{align}
Visualization of this set is shown in Fig. \ref{fig:desired_action_set_visualization} (a)(b) and Fig. \ref{fig:method_overview} (B1). We say an action-chunk, $ \mathbf A_t = [\hat{\mathbf{ a}}_t, \dots, \hat{\mathbf{ a}}_{t+T-1}]$, belongs to this set $\hat{\mathcal A}_t$ if 
$\forall i \in [0, T-1],\;
\hat{\mathbf{ a}}_{t+i} \in \hat{\mathcal A}^{\mathrm{ss}}_{t+i}.$
Action-chunks outside of this set $\hat{\mathcal A}_t$ are considered undesired. 
The information provided by this set can be used to train a policy, and we detail this in the following section.
% Desired action chunk set:
% $\hat{\mathcal{A}}_t = \{ \mathbf{A}_t =[\mathbf a_{t}, \mathbf a_{t+1},\dots, \mathbf a_{t + T_a-1} ] |  \mathbf{a}_i \in  \hat{\mathcal{A}}_i, \mathbf{a}_i^+\in\mathbf{A}^+_i,\mathbf{a}_i^-\in \mathbf{A}^-_i\}$

% \subsection{Formulation of CLIC loss with Diffusion policy}

\subsection{Policy Learning via Desired Action Sets}
\label{sec:method:policy_learning_desired_action_set}

Here, we outline the methods for training a diffusion policy using desired action sets.
% First, we extend the CLIC loss to action-chunks in Sec. \ref{sec:sub:CLIC_loss}.
First, we extend set-valued supervision from single-step actions to action-chunks in Sec. \ref{sec:sub:CLIC_loss}.
Second, to make this loss compatible with diffusion models, in Sec. \ref{sec:sub:approximation_CLIC_loss}, we detail how this loss is approximated by maximizing the likelihood of a list of desired action-chunk samples.
Finally, Sec. \ref{sec:sub:sampling_action_chunks_from_target} details the methods to generate these action-chunk samples.

\input{contents/04_02_CDP_algorithm}

% \subsubsection{CLIC Loss for Action-Chunks}
\subsubsection{Set-valued Supervision for Action-Chunks}
\label{sec:sub:CLIC_loss}

Given $[\mathbf{O}_t,\mathbf{A}^-_t, \mathbf{A}^+_t]$, our goal is to train a policy that generates action-chunks belonging to the desired action set $\hat{\mathcal{A}}_t$, defined by $\mathbf{A}^-_t $ and $ \mathbf{A}^+_t$.
Extending CLIC's set-valued supervision to action-chunks, this objective can be written as maximizing the probability that the policy samples from the desired set:
% \begin{equation}
%     \max_{ \mathbf{\theta}} \pi_{\mathbf \theta}(\mathbf{A} \in \hat{\mathcal{A}}_t  | \mathbf{O}_t)
%     \label{eq:CDP_general_objective}
% \end{equation}
\begin{equation}
    \max_{\scriptsize \mathbf{\theta}} 
    \; \pi_{\mathbf \theta}(\mathbf{A} \in \hat{\mathcal{A}}_t  | \mathbf{O}_t)
    \label{eq:CDP_general_objective}
\end{equation}
To optimize this objective, following CLIC, we adopt a policy improvement perspective by defining a target policy that assigns a higher probability to action-chunks consistent with the desired action set.
Specifically, we define a likelihood model that indicates whether an action-chunk $\mathbf{A}$ belongs to the desired action set 
\begin{align*}
    p(\mathbf{A} \in \hat{\mathcal{A}}_t | \mathbf{O}_t, \mathbf{A}) = \left\{ {\begin{array}{*{20}{c}}
	 1,&\mathbf{A} \in \hat{\mathcal{A}}_t,\\
	  0,& \text{otherwise}.
	\end{array}} \right. 
\end{align*}
Using this likelihood model to reweight the current policy yields the following target distribution:
\begin{align}
\label{eq:target_policy}
    \pi^{\text{target}}(\cdot| \mathbf{O}_t) \propto p(\mathbf{A} \in \hat{\mathcal{A}}_t | \mathbf{O}_t, \mathbf{A}) \pi_{\bm \theta}(\mathbf{A}| \mathbf{O}_t) ,
\end{align}
Intuitively, $\pi^{\text{target}}$ redistributes probability mass toward action-chunks consistent with the corrective feedback, while remaining close to the current policy.
As in CLIC, optimizing Eq.~\eqref{eq:CDP_general_objective} can be achieved by minimizing the KL divergence between the target policy and the current policy
\begin{align*}
   \ell= \mathbb{E}_{(\mathbf{A}^+, \mathbf{A}^-, \mathbf{O}_t) \sim p_\mathcal{D}} \left[ \mathrm{KL}\!\left( \pi^{\text{target}}(\mathbf{A}| \mathbf{O}_t) \big\| \pi_{\bm \theta}(\mathbf{A} | \mathbf{O}_t) \right) \! \right] 
\end{align*}
 This KL loss can be calculated for EBM-based or Gaussian-based policies in prior CLIC formulations.
 However, evaluating this loss requires access to the likelihood of an action-chunk under the policy, which is not directly available for diffusion models. 
We therefore introduce an approximation that enables optimizing this objective using diffusion models.

\subsubsection{Approximating CLIC loss}
\label{sec:sub:approximation_CLIC_loss}
The KL loss can be transformed into maximizing the likelihood of action-chunks sampled from the target policy: 
% \begin{align*}
%     \ell = - \mathbb{E}_{\mathbf{A} \sim \pi^{\text{target}}(\cdot| \mathbf{O}_t)} \left[ \pi_{\bm \theta}(\mathbf{A} | \mathbf{O}_t)  \right] 
% \end{align*}
\begin{align*}
    \ell =
    - \mathbb{E}_{\mathbf{A} \sim \pi^{\text{target}}(\cdot| \mathbf{O}_t)}
    \left[
    \log \pi_{\bm \theta}(\mathbf{A} | \mathbf{O}_t)
    \right].
\end{align*}
% Then the MLE loss can be approximated by maximizing the probability of action-chunk samples drawn from $\pi^{\text{target}}$:
We approximate the expectation with $N$ samples
% \begin{equation}
   % \mathbb{A} := \{ \mathbf{A}_{t1} , \dots,  \mathbf{A}_{tN}  \} \sim \pi^{\text{target}}(\cdot| \mathbf{O}_t)
% \end{equation}
\begin{equation}
\mathbb{A}_t := \{\mathbf{A}_{t,(i)}\}_{i=1}^{N},
\
\mathbf{A}_{t,(i)} \sim \pi^{\text{target}}(\cdot |\mathbf{O}_t)
\end{equation}
For the diffusion model, the maximum-likelihood objective reduces to the standard denoising loss \cite{2015_diffusion, 2020_diffusion}:
\begin{align}
\!\ell_t \!\!= \!\mathbb{E}_{\substack{k \sim [1\!, K]\!, \mathbf{O}_t\\ \mathbf{A}^0 \sim \mathbb{A}_t, \boldsymbol{\epsilon}_k}} 
\!\Big[\!\|\boldsymbol{\epsilon}_k \!-\!
\boldsymbol{\epsilon}_\theta(\mathbf{O}_t,\!\sqrt{\bar{\alpha}_k}\mathbf{A}^0 \!\!+\! 
\sqrt{1 \!-\! \bar{\alpha}_k}\boldsymbol{\epsilon}_k, k)\|^2 \!\Big]\!
\end{align}
This loss is the same as the BC loss of the original DP (Eq.~\eqref{eq:DP_Loss}), differing only in where the action labels come from. 
The  DP loss uses $\mathbf{A}^+_t$, whereas our SDP loss uses action-chunks sampled from $\pi^{\text{target}}$. 
Next, we describe how to generate these samples.

\subsubsection{Sampling Desired Action-Chunks}
\label{sec:sub:sampling_action_chunks_from_target}
Uniformly sampling from the desired action set does not necessarily produce samples from the target distribution $\pi^{\text{target}}$.
Although the set includes actions consistent with the correction signal, some in-set actions can still be suboptimal, and training on them may mislead the policy.
To sample action-chunks from the target distribution, we therefore adopt a constrained diffusion sampling procedure inspired by reflected diffusion methods~\cite{2023_reflected_diffusion, 2024_reflected_flow_matching}.
Concretely, we modify the standard denoising update (Eq.~\eqref{eq:DP_denoising_process}) with a boundary constraint term:
\begin{equation*}
    \mathbf{A}_t^{k-1}= \mathbf L_{\hat{\mathcal{A}}_t}\Big(\alpha \bigl(\mathbf{A}_t^{k}-\gamma \boldsymbol\epsilon_\theta(\mathbf{O}_t,\mathbf{A}_t^k,k) \bigl) + \sigma \cdot \mathcal{N} \bigl(0, I \bigl) \Big),
\end{equation*}
where $\mathbf L_{\hat{\mathcal{A}}_t}$ enforces that the intermediate samples remain within the desired set $\hat{\mathcal{A}}_t$ during denoising.
Fig.~\ref{fig:method_overview} (B2) visualizes this constrained sampling process.
Leveraging the structure of the set defined in Eq.~\eqref{eq:desired_action_set_Chunk}, the operator decomposes across individual time steps: $\mathbf L_{\hat{\mathcal{A}}_t}(\mathbf{A}_t^k) = [\mathbf L_{\hat{\mathcal{A}}_{t}^{\text{ss}}}(\mathbf{a}_{t}^k), \dots, \mathbf L_{\hat{\mathcal{A}}_{t+T-1}^{\text{ss}}}(\mathbf{a}_{t+T-1}^k)]$, where each $\mathbf{L}_{\hat{\mathcal{A}}_{t+i}^{\text{ss}}}$ operates on a single-step action:
\begin{align}
\label{eq:reflection_single_desiredA} 
\mathbf L_{\hat{\mathcal{A}}_{t+i}^{\text{ss}}}(\mathbf{a}_{t+i}^k) = \left\{ {\begin{array}{*{20}{c}}
	 \mathbf{a}_{t+i}^+,&\mathbf{a}_{t+i}^k \notin \hat{\mathcal{A}}_{t+i}^{\text{ss}},\\
	  \mathbf{a}_{t+i}^k,& \text{otherwise}.
	\end{array}} \right. 
\end{align}
In reflected diffusion, $\mathbf{L}$  reflects samples in the normal direction whenever they cross the set boundary~\cite{2023_reflected_diffusion}.
In our setting, $\hat{\mathcal{A}}_{t+i}^{\text{ss}}$ can be small, and we find that repeated normal reflections cause oscillations and increase sampling cost.
% We therefore apply a simplification: whenever a denoising step produces an action-chunk with its single-step action $\mathbf{a}_{t+i}^k$ outside $\hat{\mathcal{A}}_{t+i}^{\text{ss}}$, we directly reflect it to the corresponding positive single-step action $\mathbf{a}_{t+i}^+$.
Instead, we adopt a simplified projection strategy that directly maps actions outside $\hat{\mathcal{A}}_{t+i}^{\text{ss}}$ to $\mathbf{a}_{t+i}^+$.
We empirically validate this design choice in Sec. \ref{sec:exp:ablation_reflection}.

\subsection{Algorithm}
\label{sec:method:algorithm}

Algorithm~\ref{alg:CDP_algorithm} provides the complete procedure for SDP, including the policy update, the online interactive imitation learning (IIL) loop, and offline training.

\subsubsection{SDP policy update}
Lines~\ref{alg:line:cdp_loss}--\ref{alg:line:cdp_loss_end} shows the SDP policy update. For each contrastive action-chunk data $[\mathbf O,\mathbf A^-,\mathbf A^+]$ in a batch, we first construct a desired action set $\hat{\mathcal A}$ using Eq.~\eqref{eq:desired_action_set_Chunk} (line~\ref{alg:line:cdp_create_A}). We then sample desired action-chunks by running a truncated denoising process from step $K_A$ to $0$ (lines~\ref{alg:line:cdp_sample_A_start}--\ref{alg:line:cdp_sample_A_end}). 
To reduce the computational cost, we start denoising from an intermediate step $K_A$ rather than from the full horizon $K$. 
This process is coupled with the reflection operator $\mathbf L_{\hat{\mathcal A}}(\cdot)$ applied after every denoising step. This operator enforces per-timestep membership by reverting out-of-set actions to the corresponding positive action. In our implementation, we initialize $\mathbf A^{K_A}$ either from $\mathcal N(\mathbf 0,\mathbf I)$ or using $\mathbf A^+$; we found performance to be insensitive to this choice because of the reflection operations.
The diffusion model is then trained to imitate these sampled chunks $\mathbf{A}^0$ (lines~\ref{alg:line:cdp_Diffusion_Training_start}--\ref{alg:line:cdp_Diffusion_Training_end}).

% Unless otherwise stated, we use $K=100$ and $K_A=16$.

\subsubsection{Online learning}
Lines~\ref{alg:line:IIL_loop_start}--\ref{alg:line:IIL_loop_end} describe the online IIL loop. The robot takes actions from an action-chunk sampled from  $\pi_{\bm\theta}$, sampling a new length-$T$ chunk once the previously sampled chunk has been executed (lines~\ref{alg:line:IIL_no_human_intervention}--\ref{alg:line:IIL_execute_robot_action}). The teacher intervenes when undesired robot behavior is detected. During intervention, the human teacher's action $\mathbf a^h_t$ is executed, and we additionally query the robot policy to record the corresponding robot action $\mathbf a^r_t$ (lines~\ref{alg:line:IIL_loop_query_robot_policy}--\ref{alg:line:IIL_loop_execute_teacher_action}). We use a shorter horizon $T_r \le T$ during interventions to obtain more accurate robot actions.
Data from each step is saved to $\mathcal D_{\textnormal{traj}}$. Using a length-$T$ sliding window, once teacher actions are available for all $T$ steps, we obtain contrastive action-chunks by pairing teacher actions as positives and the corresponding robot actions as negatives, and save $[\mathbf O,\mathbf A^-\!,\mathbf A^+]$ to $\mathcal D_{+-}$ (lines~\ref{alg:line:check_A_all_actions}--\ref{alg:line:extract_contrastive}). In practice, we instruct the teacher to continue intervening for at least $T$ consecutive steps once an intervention starts, ensuring that contrastive action-chunks can be extracted.

For a given state, desired action sets constructed from feedback at different time steps can be combined, as illustrated in Fig. \ref{fig:desired_action_set_visualization}(c).
This combination refines the effective supervision region and therefore facilitates policy convergence, particularly in cases where the positive actions are noisy or suboptimal \cite{2025_CLIC}.
In practice, SDP does not explicitly combine desired action sets; rather, this refinement is approximated implicitly through policy updates from multiple corrections, and we assume that the generalization capabilities of DNNs enable
this combination across similar states. 
 Within the IIL loop, policy improvement leads to smaller desired action sets for newly collected data pairs, compared to earlier ones at the same state (Fig. \ref{fig:method_overview}).
% Within the IIL loop, the desired action sets progressively shrink as the policy improves, as shown in Fig. \ref{fig:method_overview}.

\subsubsection{Offline learning}
Lines \ref{alg:line:Offline_start}-\ref{alg:line:Offline_end} describe offline training from demonstrations or corrections.  For the demonstration dataset, each state-action pair is converted into a correction data tuple in lines \ref{alg:line:Offline_demo_label_start}-\ref{alg:line:Offline_demo_label_end}. 
Specifically, an auxiliary negative action is sampled for each $\mathbf{a}^h$  such that the resulting set $\hat{\mathcal{A}}^{\text{ss}}$ has radius $r$  (Eq.~\eqref{eq:desired_circular_space}). 
% by treating a noise-disturbed teacher action as a negative robot action.
Lines \ref{alg:line:offline_extract_correction_data}-\ref{alg:line:Offline_end} extracts the correction dataset using the same procedure as in online learning, and the policy is trained on this dataset using the SDP policy update. 

% explain how we obtain corrections during online framework

% explain how we add noise for offline demonstration

%% file: contents/04_02_CDP_algorithm.tex
\begin{algorithm*}[t!]
\caption{SDP: Set-Supervised Diffusion Policy }\label{alg:CDP_algorithm}
% \vspace{-2mm}
\begin{multicols}{2}
\DontPrintSemicolon
\SetInd{0.1em}{0.6em} %
\SetKwFunction{FDesiredA}{DesiredActionChunks}
\SetKwFunction{Flearning}{Learning}
\SetKwFunction{Fimplicit}{ImplicitPolicyShaping}
\SetKwFunction{FPolicyshaping}{PolicyShaping}
\SetKwProg{Fn}{Function}{:}{}
\SetNoFillComment
\text{Notations}
\small
\setlength{\tabcolsep}{2pt} % default is 6pt
\begin{tabular}{@{}l l}
 $\mathcal{D}_{+-}$ 
   &: Dataset of contrastive action-chunk pairs $[\mathbf{O},\mathbf{A}^- , \mathbf{A}^+]$ \\
 % $\mathcal{D}_{\textnormal{traj}}$ 
 %   & : Trajectory-level dataset, $\{[\mathbf o_t, \mathbf a^r_t, \mathbf a^ h_t] \}$ for corrections \\
 %   & \phantom{:} and $\{[\mathbf o_t,  \mathbf a^ h_t] \}$ for offline demonstrations \\
  $\mathcal{D}_{\textnormal{traj}}$ &: $\{[\mathbf o_t, \mathbf a^r_t, \mathbf a^ h_t]\}$ (corrections) or $\{[\mathbf o_t, \mathbf a^h_t]\}$ (demos) \\
 $\pi_{\boldsymbol \theta}$ 
   & : Robot policy represented via a diffusion model $\boldsymbol\epsilon_\theta$ \\
  % $\mathbf a^h_t$ &: Human intervenes if detecting undesired robot behavior\\
    $T$ &: Action-chunk length (environment execution horizon) \\
    $T_r$ &: Queried action-chunk length during intervention \\
 % $\mathbb{A}_i$ 
   % & : Set of action-chunk samples for observation $\mathbf{O}_i$ \\
    $K$ &: Total diffusion steps \\
   $K_A$
    & : Start denoising step for sampling desired action-chunks \\
 $\hat{\mathcal{A}}_i$ 
   & : Desired action set constructed from $(\mathbf A_i^+, \mathbf A_i^-)$ \\
 $b$ 
   & : In-episode update frequency \\
 $N_{\textnormal{training}}\!\!$ 
   & : End-of-episode training steps \\
  $N$ &: Number of desired action-chunk samples  \\
\end{tabular}
\normalsize

% \vspace{0.5mm}

\vspace{1mm}
\code{SDP Policy Update (Fig. \ref{fig:method_overview} (B1, B2) ) }
\label{alg:line:cdp_loss}\;
% \Fn{\FPolicyshaping{$\mathcal B$, $\{\hat{\mathbf A}\}_i$, ${\bm\theta}$}}{\label{alg:implicit_policy_shaping}
\Fn{\FPolicyshaping{$\mathcal D_{+-}$, ${\bm\theta}$} }  {\label{alg:implicit_policy_shaping}
Sample batch $\mathcal B$ from $\mathcal D_{+-}$\;
\ForEach{$[\mathbf{O},\mathbf{A}^- , \mathbf{A}^+]_i\in\mathcal B \textup{ in parallel}$ }{
Create desired action set $\hat{\mathcal A}$ from $\mathbf{A}^+ , \mathbf{A}^-$ via Eq.~\eqref{eq:desired_action_set_Chunk}
\label{alg:line:cdp_create_A}\;
% Sample $N_A$ $\mathbf{A}^{K_A} \sim \mathcal{N}(\mathbf{0}, \mathbf{I})$
Initialize $N$ samples $\mathbf{A}^{K_A}$ (e.g., $\mathcal{N}(\mathbf{0},\mathbf{I})$ or $\mathbf A^+$)
\label{alg:line:cdp_sample_A_start}\;
 \For(\tcp*[h]{\small \!\!\! Obtain samples $\mathbf{A}^0$}){$k = K_{A}, \dots, 0$}{
$\mathbf{A}^{k-1} \!\!=\alpha(\mathbf{A}^{k}-\gamma \boldsymbol\epsilon_\theta(\mathbf{O},\mathbf{A}^k,k)+ \mathcal{N} \bigl(0, \sigma^2 I \bigl))$ \;
$\mathbf{A}^{k-1} = \mathbf L_{ \hat{\mathcal{A}}}(\mathbf{A}^{k-1})$ (Eq.~\eqref{eq:reflection_single_desiredA}) 
\label{alg:line:cdp_sample_A_end}\;}
$k\sim$ Uniform$({0, \dots, K}), \epsilon_k \sim \mathcal{N}(\mathbf{0}, \mathbf{I})$
\label{alg:line:cdp_Diffusion_Training_start}\;
Take gradient descent step on \tcp*[h]{\small \!\!\! Imitate $\mathbf{A}^0$}\;
$\quad \quad \nabla_{\theta}\left[\|\boldsymbol{\epsilon}_k - \boldsymbol{\epsilon}_\theta(\mathbf{O}, \sqrt{\bar{\alpha}_k}\mathbf{\mathbf{A}}^0 + \sqrt{1 - \bar{\alpha}_k}\boldsymbol{\epsilon}_k, k)\|^2 \right]$ 
\label{alg:line:cdp_Diffusion_Training_end}\;
\Return{$\bm \theta$} \label{alg:line:cdp_loss_end}
}
}
\columnbreak
\vspace{-1mm}
\code{Interactive Imitation Learning Loop (Fig. \ref{fig:framework}A and \ref{fig:method_overview}) } \label{alg:line:IIL_loop_start} \;
\For{online episode = 1, 2, \dots }{
 \For{$t = 1, 2, \dots$}{
 Observe $\mathbf o_t$; receive $\mathbf a^h_t$ if given or set $\mathbf a^h_t \leftarrow$ None\;
 \If{\textnormal{Human intervention} $\mathbf{a}^h_{t}$ \textnormal{is None}
 \label{alg:line:IIL_no_human_intervention}
 }{
 Obtain $ \mathbf{A}_i =\mathbf{A}_t \sim \pi_{\boldsymbol \theta}( \cdot | \mathbf{O}_t)$\ if no $\mathbf{A}_{i},$  $i\in(t-T, t]$\;
 % Execute $\mathbf a^r_t$ from $\mathbf{A}_i$}
 Execute $\mathbf a^r_t =\mathbf{A}_i[t-i]$}
  \label{alg:line:IIL_execute_robot_action}
\Else(\tcp*[h]{\small \!\!\!\!Teacher \!intervenes;Query \!robot \!policy}){  
Obtain $ \mathbf{A}_i =\mathbf{A}_t \sim \pi_{\boldsymbol \theta}( \cdot | \mathbf{O}_t)$\ if no $\mathbf{A}_{i},$  $i\in(t-T_r, t]$ \label{alg:line:IIL_loop_query_robot_policy}\;
% Execute $\mathbf{a}^h_{t}$; obtain (unexecuted) $\mathbf a^r_t$ from $ \mathbf{A}_i$\label{alg:line:IIL_loop_execute_teacher_action}
Execute $\mathbf{a}^h_{t}$; obtain (unexecuted) $\mathbf a^r_t=\mathbf{A}_i[t-i]$\label{alg:line:IIL_loop_execute_teacher_action} 
}

Append $[\mathbf o_t, \mathbf a^r_t, \mathbf a^ h_t]$ to $\mathcal D_{\textnormal{traj}}$\;
% \If{$\mathbf A^ +_{t-T}$ has no None actions \label{alg:line:check_A_all_actions}} 
\If{\textnormal{None not in} $[\mathbf{a}^h_{t-T}, \dots, \mathbf{a}^h_{t-1}]$ \label{alg:line:check_A_all_actions}} 
{Get $[\mathbf O_{t-T}, \mathbf A^-_{t-T}, \mathbf A^ +_{t-T}]$ from $\mathcal D_{\textnormal{traj}}$, append  to $\mathcal D_{+-}$ \label{alg:line:extract_contrastive}\;}
% Extract $[\mathbf O_{t-T}, \mathbf A^-_{t-T}, \mathbf A^ +_{t-T}]$ from $\mathcal D_{\textnormal{traj}}$, append  to $\mathcal D_{+-}$ \label{alg:line:extract_contrastive}\;
\If{$t \% b = 0$ or $\bm a^h_t$ is provided}{ \label{alg:line:update_feq_b}
\label{alg:line:sample_batch}
      $\bm\theta \leftarrow$ \FPolicyshaping{$\mathcal D_{+-}$, ${\bm\theta}$}\;\label{alg:line:update_policy_once}
}
 }
 
 Update policy $\pi_{\bm \theta}$ as in line \ref{alg:line:update_policy_once} for $N_{\textnormal{training}}$ steps \label{alg:line:end_of_episode}\label{alg:line:IIL_loop_end}\;
}

\vspace{1mm}
\code{Offline Learning Loop (Fig. \ref{fig:framework}A and \ref{fig:method_overview}B) } \label{alg:line:Offline_start}\;
\If{$\mathcal D_{\textnormal{traj}}$\textnormal{ is Demonstration dataset} \label{alg:line:Offline_demo_label_start}}{
\ForEach{$[\mathbf{o}_t, \mathbf{a}^h_t] \in$ $\mathcal D_{\textnormal{traj}}$ }{
% $\mathbf{a}^r_t = \mathbf{a}^h_t + r \cdot \mathbf{I} / ||\mathbf{I}||$; 
$\textnormal{Sample } \mathbf{a}^r_t 
\text{  s.t. } \lVert \mathbf{a}^r_t - \mathbf{a}^h_t\rVert = 1$;
append $\mathbf{a}^r_t$ to $[\mathbf{o}_t, \mathbf{a}^h_t]$ \label{alg:line:Offline_demo_label_end}
}}
Extract $\mathcal D_{+-}$ from $\mathcal D_{\textnormal{traj}}$ as in  lines~\ref{alg:line:check_A_all_actions}--\ref{alg:line:extract_contrastive} \label{alg:line:offline_extract_correction_data}\;
\For{training episode = 1, 2, \dots}{
Update policy $\pi_{\bm \theta}$ as in line \ref{alg:line:update_policy_once} for $N_{\textnormal{training}}$ steps \label{alg:line:Offline_end}
}

\vspace*{0.05mm}
\end{multicols}
\end{algorithm*}

%% file: contents/05_Experiment.tex
\section{Experiments}
\label{sec:experiments}

We demonstrate the effectiveness of SDP through a series of simulations and real-world experiments.
In Sec. \ref{sec:exp:interactive_learning}, we compare SDP with state-of-the-art methods within the online IIL framework. 
In Sec. \ref{sec:exp:offline_learning_results}, we assess the quality of datasets collected by SDP and demonstrate that SDP can be trained effectively from offline datasets.
In Sec. \ref{sec:exp:ablation}, ablation studies are carried out to study the effectiveness of the key components of SDP.
In Sec. \ref{sec:exp:real_world_results}, we showcase the effectiveness of SDP in real-world experiments. 
Detailed experimental settings are provided in the Appendix.

% explain the setup; baseline
\textbf{Baselines}: We compare our method against DP \cite{2023_diffusionpolicy}, DP-DPO \cite{2024cvpr_Diffusion_DPO}, Ambient DP (ADP) \cite{2023ambient_DP}, CLIC \cite{2025_CLIC}, and Implicit Behavior Cloning (IBC) \cite{2022_implicit_BC}.
Our SDP shares the same diffusion policy structure as DP, DP-DPO, and ADP, differing only in the loss function.
DP is trained with a standard BC loss; DP-DPO adds additional direct preference optimization loss to DP; ADP extends DP to recover clean action distributions from highly-corrupted action data.
CLIC and IBC use EBMs as policy representation and output single-step actions.

\textbf{Tasks}:
We compared these methods across four simulated tasks, including a Push-T task  \cite{2023_diffusionpolicy} and three manipulation tasks from the robosuite benchmark \citep{2020_robosuite} (see Appendix for task details).
The observation includes images and the robot's proprioceptive states.
For interactive learning experiments, all the agents use absolute positions as actions.  Both accurate and Gaussian-noise-corrupted intervention data are considered.
Gaussian noise is added to the intervention actions to evaluate the robustness of each method to noisy action data.
For offline learning, we evaluate both position and velocity controllers with action-chunk horizons $T=1$ and $T=16$.

\textbf{Dataset sources}: 
In online learning experiments, the data $\mathcal{D_{\text{traj}}}$ are collected during the online interactions, following lines \ref{alg:line:IIL_loop_start}-\ref{alg:line:IIL_loop_end} in Algorithm \ref{alg:CDP_algorithm}. 
The agents learn from a simulated teacher to ensure repeatability. 
In offline learning, we consider three dataset types:
(1) the dataset accumulated during the interactive learning of the SDP agent; (2) the dataset collected by DP under the same online framework; (3) the Robomimic dataset \cite{2022_robomimic}.
For both velocity and position control, SDP and DP datasets are collected with chunk length $T=16$ and subsampled to contain the same number of contrastive action-chunk pairs. 
From the original Robomimic dataset, we construct separate datasets corresponding to each controller type.
All offline experiments are conducted on the Square task.

\subsection{Results of Online Interactive Learning}
\label{sec:exp:interactive_learning}

\input{tables/sim_exps_1}

Table \ref{tab:sim_exp_interactive} summarizes the results of interactive learning. 

\subsubsection{SDP better utilizes contrastive action data than DP-DPO and CLIC}

Both SDP, DP-DPO, and CLIC leverage contrastive action data by incorporating positive and negative action information. However, their performance differs substantially, indicating different abilities to exploit contrastive action data effectively (Table~\ref{tab:sim_exp_interactive}).
DP-DPO performs worse than DP, suggesting that its pairwise comparison loss is less effective than the BC loss in manipulation tasks with the action-chunk space.
One possible explanation is that the pairwise comparison objective provides a relatively weak supervision signal: it constrains only the relative ordering between two action-chunks,
leaving the rest of the action-chunk space largely unconstrained.
This limitation becomes more pronounced in higher-dimensional or more challenging tasks: DP-DPO performs close to DP on Push-T and PickCan, but degrades substantially on Square and TwoArmLift.
In contrast, SDP outperforms DP-DPO, demonstrating that desired-action-set supervision is more effective at leveraging correction data than the pairwise comparison objective.
SDP also outperforms CLIC, suggesting the advantage of diffusion models over EBMs in modeling high-dimensional action-chunk data. 

% \subsubsection{SDP outperforms DP counterparts through the utilization of negative actions}

\subsubsection{SDP outperforms DP counterparts through the set-valued supervision}

Across both accurate and noisy feedback settings, SDP consistently outperforms DP (Table~\ref{tab:sim_exp_interactive}), highlighting the benefit of explicitly incorporating negative actions during learning.
DP directly imitates feedback actions using a BC objective, making its performance strongly dependent on data quality. As feedback transitions from accurate to noisy, DP experiences significant performance degradation due to the direct influence of corrupted labels.
In contrast, SDP trains diffusion policies with desired action sets rather than exact action labels, allowing the learned actions to deviate from the action label and therefore enabling SDP to improve the policy even under noisy data.
This results in substantially smaller performance drops under noisy conditions.
 A similar trend is observed between CLIC and IBC: IBC degrades more significantly due to its reliance on exact imitation.
ADP has access to privileged information in the form of optimal actions to obtain the corruption matrix, which allows it to outperform DP under noisy conditions. However, even with this advantage, ADP underperforms SDP, underscoring the robustness of SDP to noisy intervention feedback.

\subsection{Results of offline learning from different datasets}
\label{sec:exp:offline_learning_results}
 
% This section evaluates offline policy training.
% Results are reported in Fig. \ref{fig:exp_offline_training_simulation}. 

\begin{figure}[t!]
	\centering
	\includegraphics[width=0.49\textwidth]{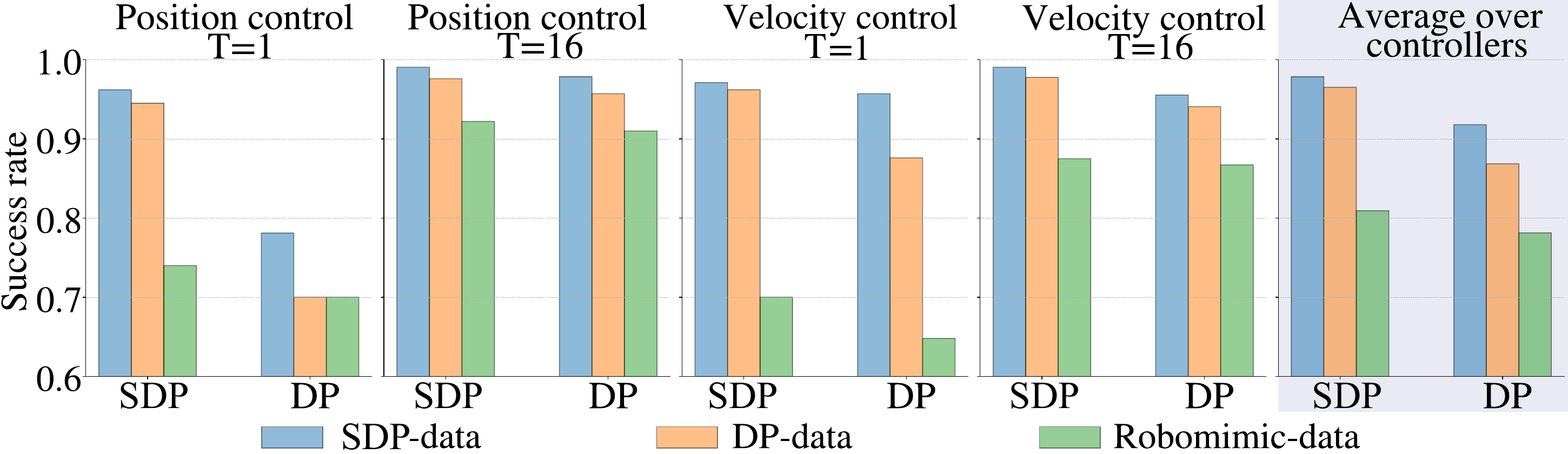}
	\caption{
 Results of offline training in the Square task.
 }
\label{fig:exp_offline_training_simulation}
\end{figure}

 Fig. \ref{fig:exp_offline_training_simulation} summarizes the results of offline policy training.

\subsubsection{SDP generates higher-quality datasets during online data aggregation} 
\begin{figure}[t!]
	\centering
	\includegraphics[width=0.475\textwidth]{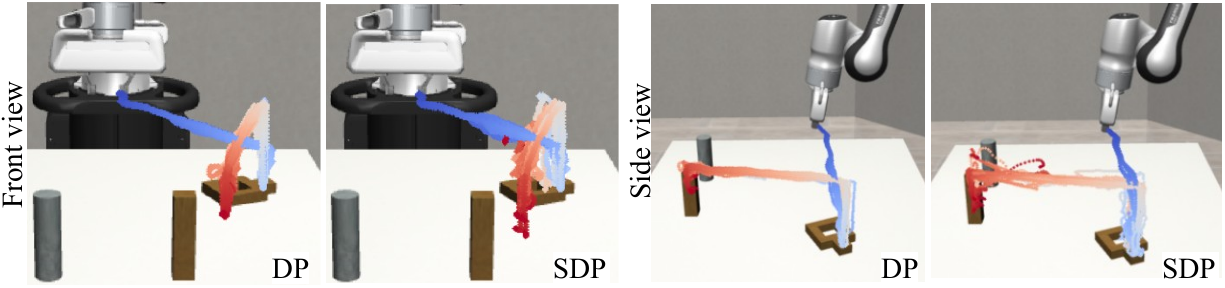}
	\caption{
Visualization of the data obtained during one interactive learning using DP and SDP agents for the Square task. 
Initial states of all episodes are fixed for clear visualization.
Only the position of the robot's end effector is visualized. 
The trajectories of these position data are projected onto two cameras. 
DP has a narrower state coverage that follows the teacher policy, while SDP explores by taking actions from the desired action set, creating a broader state coverage. 
 }
\label{fig:traj_visualization}
\end{figure}

For both SDP and DP policies, models trained on data collected by SDP consistently outperform those trained on DP-generated datasets. This performance gap suggests that SDP produces datasets with more favorable properties for offline learning.
To understand this difference, we visualize the collected trajectories in Fig.~\ref{fig:traj_visualization}. Compared to DP, SDP produces data with broader coverage of the state space.
This stems from a key difference in online data collection: DP-based DAgger strictly imitates the teacher’s action with the BC objective, whereas SDP uses set-valued supervision and allows the policy to output actions from the desired action set.
During early episodes of the online learning process, although actions sampled from the desired action set may be suboptimal, they encourage exploration around teacher-induced trajectories rather than strictly following them.
This exploration behavior emerges from the set-valued supervision used by SDP and yields a dataset with broader coverage, thereby improving the quality of the aggregated dataset.

\subsubsection{SDP outperforms DP across all datasets}

Across all evaluated datasets, policies trained with offline SDP consistently outperform their DP counterparts.
The performance gap is most significant under the controller using small action-chunks (T=1), which are less resilient to distributional shift than large action-chunks.
DP also performs worse with velocity control.
In contrast, SDP maintains superior performance across all controllers using online-collected datasets. 
We conjecture that SDP uses desired action sets to induce smoother action labels, which may reduce overfitting to individual action-chunks in the dataset and improve generalization to states near the training distribution.
SDP also outperforms DP when trained on the Robomimic dataset, demonstrating that set-valued supervision can be applied to standard offline demonstration datasets, even in the absence of negative action-chunks.

\subsection{Ablation study}
\label{sec:exp:ablation}

\label{sec:exp:ablation_reflection}
SDP demonstrates stable performance across a wide range of radius ratios for desired action sets, with ablations reported in the Appendix.
Here, we study how different sampling strategies affect the quality of sampled desired action-chunks.  
Once a desired action set is defined from the correction data, SDP requires a practical sampling method that generates action-chunks within this set while remaining close to the current policy distribution, as described in Sec.~\ref{sec:sub:sampling_action_chunks_from_target}.
We therefore evaluate our reflected sampling method against four alternatives within the online interactive learning framework: (1) \textbf{Random reflection}, which follows our reflection procedure but projects out-of-set samples to random points inside the set; (2) \textbf{Classifier guidance}, which treats the desired action set as a classifier and biases the denoising process toward $\mathbf{a}^+$; (3) \textbf{Classifier guidance + final filter}, which removes out-of-set samples after guidance; and (4) \textbf{Classifier guidance + final reflection}, which applies a single reflection step at the final denoising iteration following classifier guidance. Results are reported in Table~\ref{tab:ablation_reflection}.

\begin{table}[h]
\centering
\setlength{\tabcolsep}{4pt} % reduce column spacing
\renewcommand{\arraystretch}{1.1} % slightly tighter rows
\caption{Ablations on sampling desired action-chunks}
\label{tab:ablation_reflection}
\begin{tabular}{lccccc}
\Xhline{0.75pt}
& Classifier 
& Guidance+ 
& Guidance+ 
& Reflection 
& Random \\
 & guidance & final filter & final reflection & to $\mathbf{a}^+$ & reflection \\
\hline
Accurate & 0.890  & 0.429 & 0.964 &  \textbf{0.988}  &  \textbf{0.978} \\
Noisy    & 0.448 & 0.452 & 0.771 &  \textbf{0.946} &  \textbf{0.933} \\
\Xhline{0.75pt}
\end{tabular}
\label{tab:gs_reflection_results}
\end{table}

Our default \textbf{Reflection to $\mathbf{a}^+$} achieves consistently strong performance and matches \textbf{Random reflection}, indicating that enforcing set membership via reflection is more important than the exact projection strategy.
In contrast, \textbf{Classifier Guidance} performs substantially worse.
While adding a final reflection step improves its performance, it still lags behind reflection-based sampling in the noisy setting.
Overall, these results show that classifier guidance fails to reliably produce in-set desired action-chunks, whereas reflection-based sampling can enforce the set constraint while maintaining sample quality.

\subsection{Results of real-world Experiments}
\label{sec:exp:real_world_results}

We evaluate the effectiveness of SDP on two challenging real-world, long-horizon manipulation tasks: a multi-modal Insert-T task \cite{2025_CLIC} and a roundtable assembly task from FurnitureBench \cite{2023RSS_2025IJRR_furniture_bench}.
For comparison, we also evaluate DP, trained using the same datasets as SDP.
The policy observations include RGB images from two cameras and the robot's end-effector pose.
The experiments are carried out using a 7-DoF KUKA iiwa manipulator for the Insert-T task and a 7-DoF Franka Panda manipulator for the roundtable task. 
 A 6D space mouse is employed to provide intervention feedback on the pose of the robot's end effector. Furthermore, in the roundtable task, a keyboard provides feedback on the gripper's actuation.

%1. introduce how we collect data (demonstration + correction)

\subsubsection{Insert-T task}
\input{tables/real_exp_table1}

\begin{figure}[t!]
	\centering
	\includegraphics[width=0.455\textwidth]{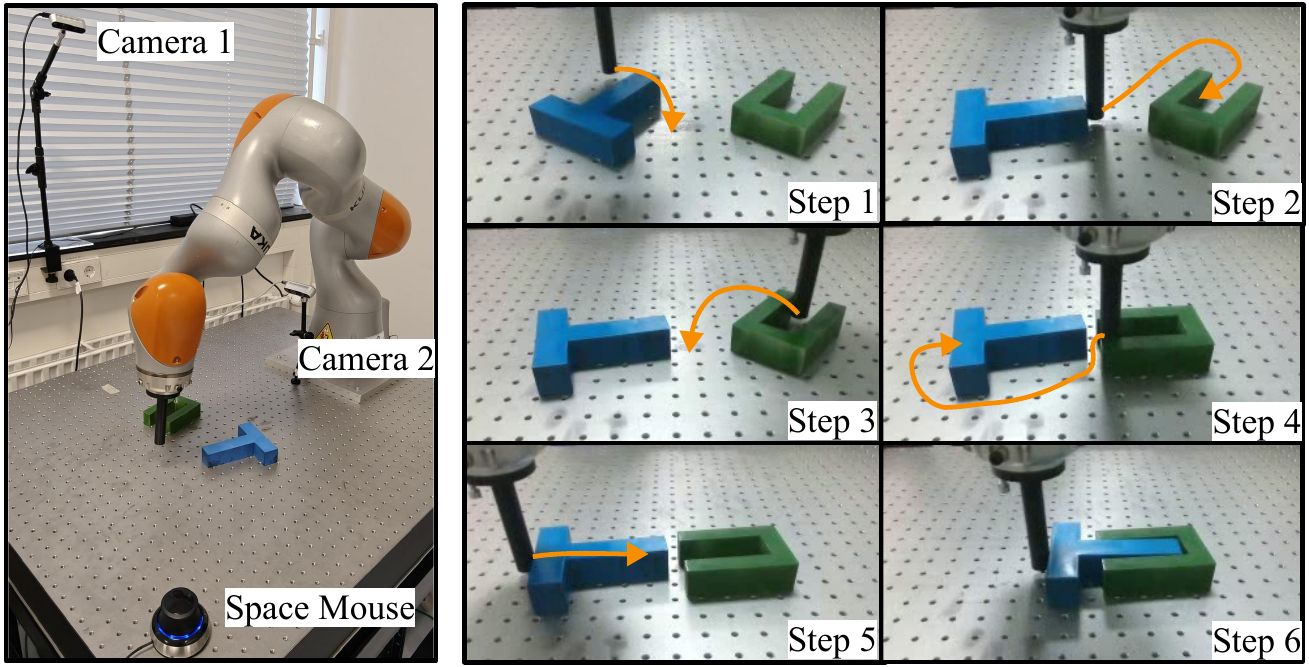}
	\caption{
Left: Hardware setup for the Insert-T task. Right: An example of SDP policy rollout during evaluation (cropped views from camera 2).
 }
\label{fig:InsertT_exp}
\end{figure}

%1. introduce the task 
%1.2 introduce how the dataset are collected, and how you train/evaluate the model
The Insert-T task requires the robot to insert a T-shaped object into a U-shaped object by pushing to adjust their positions and orientations.
The setup is shown in Fig. \ref{fig:InsertT_exp}.
Three difficulty levels can be defined \cite{2025_CLIC}—easy ($<$1 contact changes), medium ($<$5), and hard ($\geq$5)—based on the required contact changes from a teacher policy.
We first collect 50 demonstrations to pretrain the policy using each method. 
Then the pretrained SDP policy is deployed on the robot, during which a human teacher provides corrective interventions, resulting in 40 episodes of intervention data.
Each method is subsequently trained on the combined dataset of demonstrations and corrections.
For both the hard and medium tasks, we evaluate the methods with the same set of 40 initial states. The results are reported in Table \ref{tab:insert}.

%2. analyse the results and draw the key finding
SDP consistently outperforms DP in terms of success rate, both after pretraining and when trained on the full dataset. 
These results align with the offline learning results in Sec. \ref{sec:exp:offline_learning_results}.
While both methods benefit from the correction dataset, SDP exhibits substantially larger improvements on the hard task.
This suggests that SDP is more effective at leveraging intervention data, as it can leverage negative actions, whereas DP can only imitate human actions.

\begin{figure}[t!]
	\centering
	\includegraphics[width=0.48\textwidth]{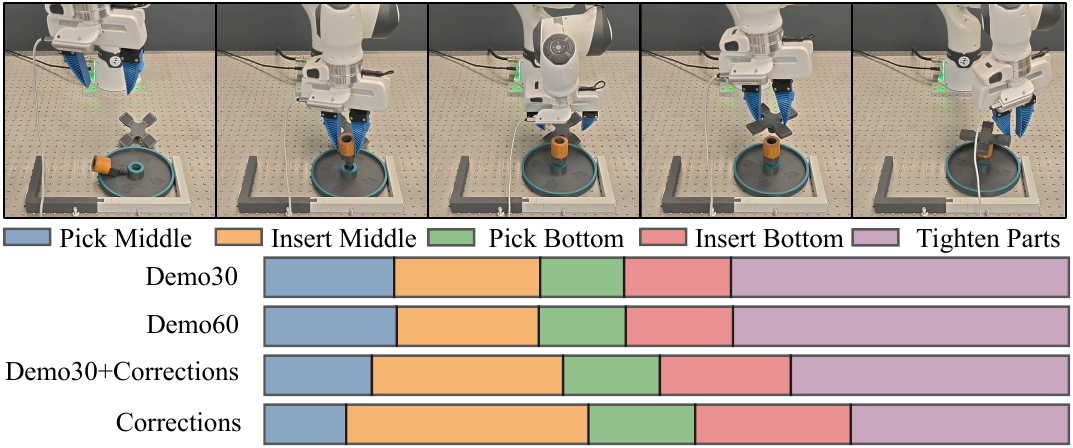}
	\caption{\!\!\!\;Stage distribution across datasets in the round-table task.\!\!}
\label{fig:dataset_stage_distribution_roundtable}
\end{figure}
\begin{figure*}[t!]
   \setlength{\abovecaptionskip}{3pt}
	\centering
	\includegraphics[width=0.97\textwidth]{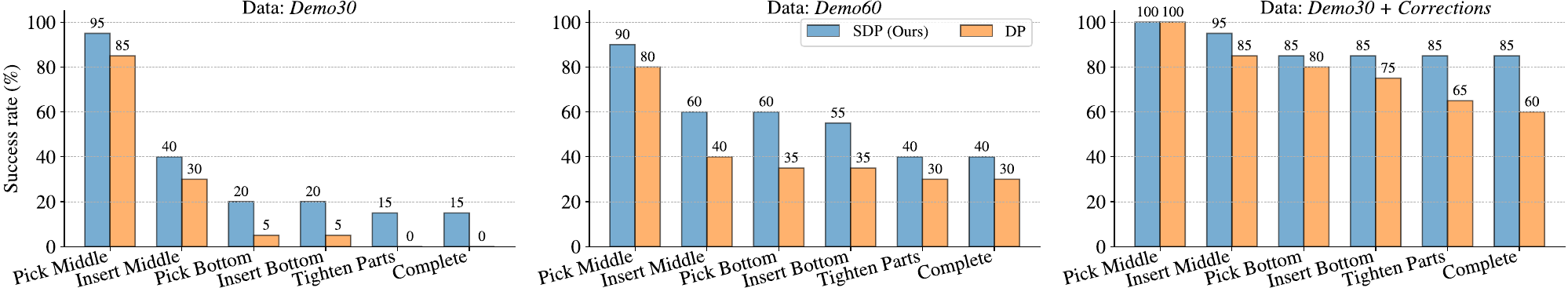}
	\caption{
Round-table assembly results with stage-wise success breakdown. For each dataset and method, we report the fraction of 20 trials that successfully progress to each stage of the task sequence; “Complete” corresponds to overall task success. 
% CDP and DP are trained on Demo30, Demo60, and Demo30 + Corrections.
}
\label{fig:roundtable_result}
\end{figure*}

\subsubsection{Round-table Assembly task}
The round-table assembly task \cite{2023RSS_2025IJRR_furniture_bench} requires the robot to perform a sequence of stages, as shown in the upper figure of Fig. \ref{fig:dataset_stage_distribution_roundtable}.
Such long-horizon tasks are challenging for offline IL methods \cite{2025_IL_Recovery_Correction} as even slight differences in the dataset at a given stage can compound and lead to task failure.
We first collect 30 demonstrations (30,626 observation-action pairs) to pretrain both SDP and DP. 
Then, we deploy the SDP policy, and a human teacher provides corrective interventions on its undesired behavior. 
This deployment-and-correction process is repeated for three iterations, producing a final correction dataset consisting of 29,194 contrastive action-chunk pairs.
Additionally, to assess the data quality of the correction dataset, we collected 30 extra demonstration datasets.
This results in three training datasets: \textit{Demo30} (initial 30 demonstrations), \textit{Demo30 + Corrections} (30 demonstrations plus corrective interventions), and \textit{Demo60} (60 demonstrations).
% We therefore have three datasets to train each method:  (1) the initial 30 demonstrations (Demo 30); (2) the initial 30 demonstrations and the correction dataset (Demo 30 + corrections); (3) the combined 60 demonstrations (Demo 60). 
Both SDP and DP are trained on \textit{Demo30} for 12 hours with an Nvidia A40 GPU, after which each checkpoint is further trained for an additional 12 hours using either \textit{Demo30 + Corrections} or \textit{Demo60}.
For the methods trained with each dataset,
we evaluate the last checkpoint with the same 20 initial states. The results are reported in Fig. \ref{fig:roundtable_result}.

Both SDP and DP achieve higher success rates with \textit{Demo30 + Corrections} than with \textit{Demo60} (with roughly the same amount of data). 
As illustrated in Fig. \ref{fig:dataset_stage_distribution_roundtable}, demonstration data is strongly biased toward stages with longer execution time, such as the `Tighten Parts' stage, which accounts for 41\% of the total steps but is not the primary bottleneck stage.
In contrast, correction data is naturally concentrated around failure-prone bottleneck stages, such as `Insert Middle' and `Insert Bottom'.
% , and includes recovery behaviors that guide the policy back toward the data distribution.
This explains the improved performance observed when corrective interventions are incorporated.

SDP consistently outperforms DP across all three datasets. When trained on \textit{Demo30} and \textit{Demo60}, SDP achieves approximately 10\% higher success rates, as SDP can imitate the behavior of the dataset while avoiding overfitting to each individual data point.
Notably, when trained with the \textit{Demo30 + Corrections} dataset, the performance gap between SDP and DP is the largest.
We attribute this result to SDP's ability to utilize the correction dataset, whereas DP only imitates the positive action-chunk and discards the negative ones. 

Together with the Insert-T results, these findings indicate that SDP not only improves overall success rates in long-horizon manipulation tasks, but also makes more effective use of human corrective feedback than DP.

% \cite{2024iros_juicer_IL_furniture, 2025icra_IL_RL_refinement_furniture}

% \input{tables/real_exp_furniture}

%% file: tables/sim_exps_1.tex
\begin{table}[t!]
\footnotesize
\setlength{\tabcolsep}{4pt}
\caption{Success rates of interactive learning experiments in simulation under accurate and noisy intervention feedback.
}
\label{tab:sim_exp_interactive}
\begin{center}
\begin{tabular}{lcccccc}
\Xhline{0.75pt}
  & SDP & DP\cite{2023_diffusionpolicy} & \!DP-DPO\cite{2024cvpr_Diffusion_DPO}  \!\!\! & \!ADP\cite{2023ambient_DP} & CLIC\cite{2025_CLIC} & IBC\cite{2022_implicit_BC}   \\
 \hline
 \textbf{Accurate} &  &  &  &  &  & \\
Push-T & \textbf{0.729} & 0.633  & 0.633  & 0.600 & 0.527 &  0.486\\
 Square & \textbf{0.988}  & 0.951 & 0.689 & 0.871 & 0.809 &  0.638 \\
 PickCan & \textbf{0.998} & \textbf{0.993} & 0.924  &  \textbf{0.996} & 0.962 &  0.897 \\
 TwoArmLift & \textbf{0.957}  & 0.934 & 0.518 & 0.859 &  0.789 & 0.214 \\
 \hline
Average & \textbf{0.918} & 0.878 & 0.691 & 0.832 & 0.772 & 0.559 \\
 \hline
 \textbf{Noisy} &  &  &  &  &  & \\
Push-T & \textbf{0.600} & 0.333  &0.400  &  0.410 & 0.390 & 0.333 \\
 Square & \textbf{0.946} & 0.683 & 0.623 &  0.838&  0.429 &  0.000\\
 PickCan &  \textbf{0.990}  & 0.933 & 0.933  & \textbf{0.995} & 0.900 &  0.838\\
 TwoArmLift &\textbf{0.924} & 0.895  & 0.529 & 0.881 & 0.705 & 0.000 \\
 \hline
Average & \textbf{0.865} & 0.711 & 0.621 & 0.781 & 0.606 & 0.293 \\
\Xhline{0.75pt} 
\end{tabular}
\end{center}
\end{table}

%% file: tables/real_exp_table1.tex
\begin{table}[h!]
\setlength{\tabcolsep}{6pt}
\centering
\caption{Results of offline learning in the Insert-T task.
% For calculating CT, $\diagdown$ entries are replaced with the maximum allowable timestep.
}
\label{tab:insert}
\begin{tabular}{l|cc|cc}
\Xhline{0.75pt}
 & \multicolumn{2}{c|}{Pretraining with Demonstration} & \multicolumn{2}{c}{Demonstration + Correction} \\
\cline{2-5}
 & Hard & Medium & Hard & Medium \\
\hline
SDP & 0/40 & \textbf{23/40} & \textbf{35/40} &  \textbf{39/40} \\
DP & 0/40 & 18/40 & 23/40 &  35/40\\
% DP-DPO (?) &  &  &  &  \\
\Xhline{0.75pt}
\end{tabular}
\end{table}

%% file: contents/06_Conclusion.tex
\section{Conclusion} 
\label{sec:conclusion}

% In this work, by introducing SDP, we addressed a key inefficiency in diffusion-based learning from human interventions: the negative robot's actions are ignored.

In this work, we introduced SDP, a set-supervised framework for training diffusion policies from human corrections.
Instead of treating each teacher's action as an exact action target, SDP uses the correction data to construct set-valued action targets.
To achieve this, SDP extends the desired-action-set formulation of CLIC to the action-chunk space and proposes a training pipeline that optimizes diffusion policies with these sets.
Extensive experiments in both offline and online settings demonstrate consistent performance gains and improved data quality.

\textbf{Limitations and Future work:} Our work assumes intervention feedback and thus cannot be applied to relative corrections \cite{2025_CLIC}, which might break the approximation in Sec. \ref{sec:method:desired_action_set}.
Additionally, SDP requires sampling desired action-chunks during training, 
which takes around $38\%$ of the total training time in our experiment and therefore introduces an additional computational cost.
Future work could explore extending SDP to alternative feedback modalities and improving sampling efficiency during training.

%% file: contents/07_Appendix.tex
\section{Appendix}

This appendix supplements the main paper with additional analyses and implementation details. Appendix \ref{appendix:ablation_radius} presents ablation results on the radius ratio, while Appendix \ref{Appendix:computation_cost_sampling_DesiredA} analyzes the computational cost of sampling desired action chunks. Policy implementation details are provided in Appendix \ref{Appendix:policy_implementation_details}. Appendix \ref{Appendix_details_simulation} details the simulation experiments, and Appendix \ref{Appendix:details_of_real_world} reports details of the real-world experimental setup and results.

\subsection{Ablation results for the radius ratio of desired action sets}
\label{appendix:ablation_radius}
\begin{figure}[h!]
	\centering
	\includegraphics[width=0.39\textwidth]{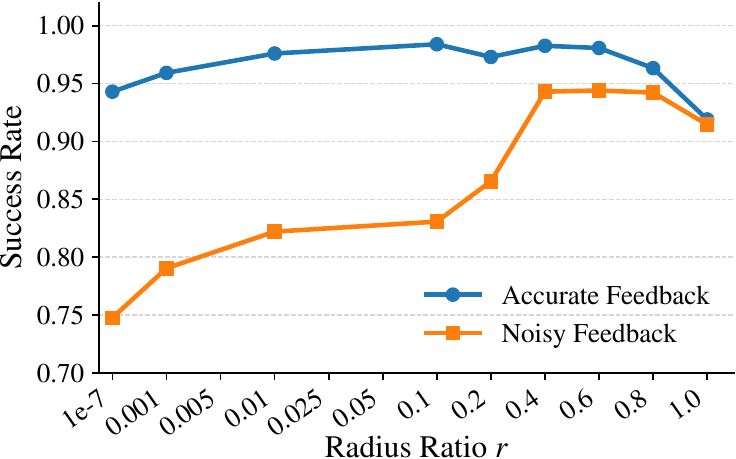}
	\caption{
Effects of $r$ in the online learning setting.
 }
\label{fig:Appendix_cdp_radius_ablation}
\end{figure}

\begin{figure}[h!]
	\centering
	\includegraphics[width=0.4\textwidth]{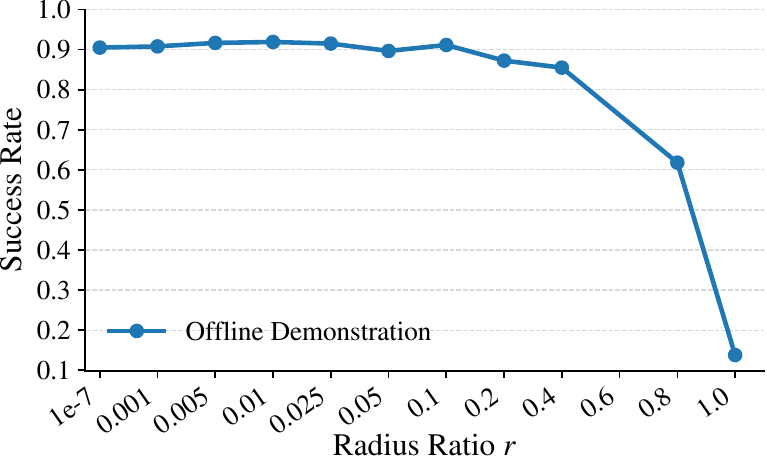}
	\caption{
Effects of $r$ in the offline learning setting with position control and T=16, using the Proficient-Human (PH) dataset in Robomimic \cite{2022_robomimic}.
The radius of each single-step desired action set is equal to $r$, as we enforce the auxiliary negative actions to satisfy $\lVert \mathbf{a}^r_t - \mathbf{a}^h_t\rVert = 1$ in \ref{alg:line:Offline_demo_label_start}-\ref{alg:line:Offline_demo_label_end} of Algorithm \ref{alg:CDP_algorithm}.
 }
\label{fig:Appendix_offline_online_radius_ablation}
\end{figure}

We evaluate SDP across a wide range of radius ratios (Eq.~\eqref{eq:desired_circular_space}). As shown in Fig.~\ref{fig:Appendix_cdp_radius_ablation} and Fig.~\ref{fig:Appendix_offline_online_radius_ablation}, SDP maintains stable and consistently strong performance across both online and offline learning settings, indicating low sensitivity to this hyperparameter in practice. 
For data with accurate positive actions, SDP's performance degrades only when the radius is excessively large, where the desired set no longer provides sufficiently informative supervision.
For data with noisy positive actions, overly small values of $r$ can lead to worse performance by restricting the desired set to a narrow region that may exclude optimal actions. 
In the limiting case with the radius being $1e-7$, the desired action set reduces to the positive action-chunk $\{\mathbf{A}^+\}$, and SDP's performance also becomes similar to DP's.

\subsection{Computation cost for sampling desired action-chunks}
\label{Appendix:computation_cost_sampling_DesiredA}
At each training step, given a batch size $B$ and $K$ sampled action chunks per data pair, we sample action-chunks of shape ($B\cdot K$, $T$, action dim) = (64 * 16, 16, 10) (line \ref{alg:line:cdp_sample_A_start}-\ref{alg:line:cdp_sample_A_end}).
On an NVIDIA A40 GPU, this sampling procedure takes approximately 0.233 seconds per training step.
For offline learning with 296 episodes and 200 training steps per episode, the total training and evaluation process lasts around 10 hours. 
Within the total time, sampling desired action-chunks takes around 3.8 hours and takes around 38$\%$ of the total time.

\subsection{SDP Policy Update with Cached Desired Samples}

To reduce the computational cost of repeatedly sampling desired action-chunks, we introduce a variant of SDP policy update that maintains a double-ended queue (deque) of previously sampled desired chunks for each contrastive action-chunk pair.
At each policy update, only $N_q$ new samples are generated through constrained denoising and added to the deque.
The deque keeps at most $N$ recent samples, which are reused as training targets for the diffusion denoising loss.
This cached-sample variant is presented in Algorithm~\ref{alg:CDP_algorithm_deque}. Lines related to the cached-sample deque are highlighted in blue.

Anecdotally, moderate refresh rates, such as $N_q\in\{2,4,8\}$ with $N=16$ or $N_q=4$ with $N=32$, produced performance close to the default SDP update in our runs on the Square task within the online IIL setting.
In contrast, very small refresh ratios $N_q/N$, such as $N_q=1,N=16$ or $N_q=2,N=32$, tended to reduce performance.
These observations suggest that refreshing the deque too slowly can cause cached samples to become less aligned with the current policy distribution, potentially misleading the policy update.

\begin{algorithm}[h!]
\caption{SDP Policy Update with Sample Deque }\label{alg:CDP_algorithm_deque}

\definecolor{dequeblue}{RGB}{0,80,160}
\newcommand{\deque}[1]{\textcolor{dequeblue}{#1}}
\DontPrintSemicolon
\SetInd{0.1em}{0.6em} %
\SetKwFunction{FPolicyshapingDeque}{PolicyShapingDeque}
% \SetKwFunction{Flearning}{Learning}
% \SetKwFunction{Fimplicit}{ImplicitPolicyShaping}
% \SetKwFunction{FPolicyshaping}{PolicyShaping}
\SetKwProg{Fn}{Function}{:}{}
\SetNoFillComment
\text{Notations}
\small
\setlength{\tabcolsep}{2pt} % default is 6pt
\begin{tabular}{@{}l l}
 $\mathcal Q_i$ 
   &: Desired-sample deque for $[\mathbf{O},\mathbf{A}^- , \mathbf{A}^+]_i$ \\
  $N_q$ &: Number of new desired samples per update\\
  $N$ &: Maximum deque size / number of cached samples  \\
\end{tabular}
\normalsize
% \code{SDP Policy Update with Sample Deque}
\label{alg:line:sdp_loss_deque}\;
\Fn{\FPolicyshapingDeque{$\mathcal D_{+-}$, ${\bm\theta}$} }  {\label{alg:sdp_policy_shaping_deque}
Sample batch $\mathcal B$ from $\mathcal D_{+-}$\;
\ForEach{$[\mathbf{O},\mathbf{A}^- , \mathbf{A}^+]_i\in\mathcal B \textup{ in parallel}$ }{
Create set $\hat{\mathcal A}$ from $\mathbf{A}^+ , \mathbf{A}^-$ via Eq.~\eqref{eq:desired_action_set_Chunk}
\label{alg:line:sdp_deque_create_A}\;

\deque{Retrieve or initialize sample deque $\mathcal Q_i$} \;
% Initialize $\mathcal Q_i$ as an empty deque if it does not exist\;
\deque{Set $N_i \leftarrow N$ if $|\mathcal Q_i| < N$, otherwise $N_i \leftarrow N_q$}\;
Initialize $N_i$ samples $\mathbf{A}^{K_A}$ (e.g., $\mathcal{N}(\mathbf{0},\mathbf{I})$ or $\mathbf A^+$)
\label{alg:line:sdp_deque_sample_A_start}\;

\For(\tcp*[h]{\small obtain samples $\mathbf{A}^0$}){$k = K_{A}, \dots, 0$}{
$\mathbf{A}^{k-1} \!\!=\alpha(\mathbf{A}^{k}-\gamma \boldsymbol\epsilon_\theta(\mathbf{O},\mathbf{A}^k,k)+ \mathcal{N} \bigl(0, \sigma^2 I \bigl))$ \;
$\mathbf{A}^{k-1} = \mathbf L_{ \hat{\mathcal{A}}}(\mathbf{A}^{k-1})$ (Eq.~\eqref{eq:reflection_single_desiredA}) 
\label{alg:line:sdp_deque_sample_A_end}\;
}
\deque{Update $\!\mathcal Q_i\!$ with $\!\mathbf{A}^{0}\!$ and keep the recent $\!N$ samples}\;

\deque{Obtain $\tilde{\mathbf{A}}^0$ from $\mathcal Q_i$
\tcp*[h]{\small \!\!\!reuse $\!\!N\!\!$ cached \!samples}}\;
$k\sim$ Uniform$({0, \dots, K}), \epsilon_k \sim \mathcal{N}(\mathbf{0}, \mathbf{I})$
\label{alg:line:sdp_deque_Diffusion_Training_start}\;
Take gradient descent step on \tcp*[h]{\small imitate $\tilde{\mathbf{A}}^0$}\;
$\quad \quad \nabla_{\theta}\left[\|\boldsymbol{\epsilon}_k - \boldsymbol{\epsilon}_\theta(\mathbf{O}, \sqrt{\bar{\alpha}_k}\tilde{\mathbf{A}}^0 + \sqrt{1 - \bar{\alpha}_k}\boldsymbol{\epsilon}_k, k)\|^2 \right]$ 
\label{alg:line:sdp_deque_Diffusion_Training_end}\;
}
\Return{$\bm \theta$} \label{alg:line:sdp_deque_loss_end}
}
\end{algorithm}
\subsection{Policy Implementation Details}
\label{Appendix:policy_implementation_details}
\subsubsection{Observation encoder}
\label{appendix:obs_encoder}
For simulation experiments, we adopt the same CNN-based observation encoder from Diffusion Policy \cite{2023_diffusionpolicy}.
Each image observation is processed by a ResNet-18 backbone, followed by spatial softmax pooling to produce a compact latent representation.

For real-world experiments, we modify the encoder to improve robustness under visual variability.
Concretely, we replace spatial softmax pooling with adaptive average pooling, which yields a 256-dimensional latent embedding.
To further regularize the learned representation, we attach an image decoder that reconstructs the original observation from the latent embedding.
The encoder is shared between the policy and the image decoder, and the reconstruction objective acts as an auxiliary loss during policy learning.
In addition, we recorded an extra free-play dataset to train this autoencoder (see Appendix \ref{appendix:data_collection}).

\subsubsection{Policy decoder}
We use the UNet-based diffusion model from Diffusion Policy \cite{2023_diffusionpolicy}.
The latent observation embedding is injected via FiLM conditioning at each denoising block.
During training, we employ a DDPM scheduler with $K = 100$ diffusion steps.
During inference, to reduce runtime, we adopt a DDIM scheduler.
We use 16 denoising steps for all tasks, except TwoArmLift, which uses 32 steps due to its higher-dimensional action space.

\subsubsection{Hyperparameters}

\input{tables/appendix_hyperparameter}

Hyperparameters used for SDP are summarized in Table \ref{tab:appendix_hyperparameters}.
All experiments use a batch size of 64 and an initial learning rate of 0.002.
Regarding the learning-rate scheduling, we use a cosine schedule with a linear warmup.
During online learning, the in-episode update frequency $b$ is fixed to 2. 

For diffusion-based baselines, the network structure and hyperparameters are the same, except that $N_{\text{training}}=1000$ for baselines, which results in a similar wall-clock training time as SDP ($N_{\text{training}}=200$) on an Nvidia A40 GPU.

\subsection{Details of Simulation Experiments}
\label{Appendix_details_simulation}

\begin{figure}[h!]
	\centering
	\includegraphics[width=0.49\textwidth]{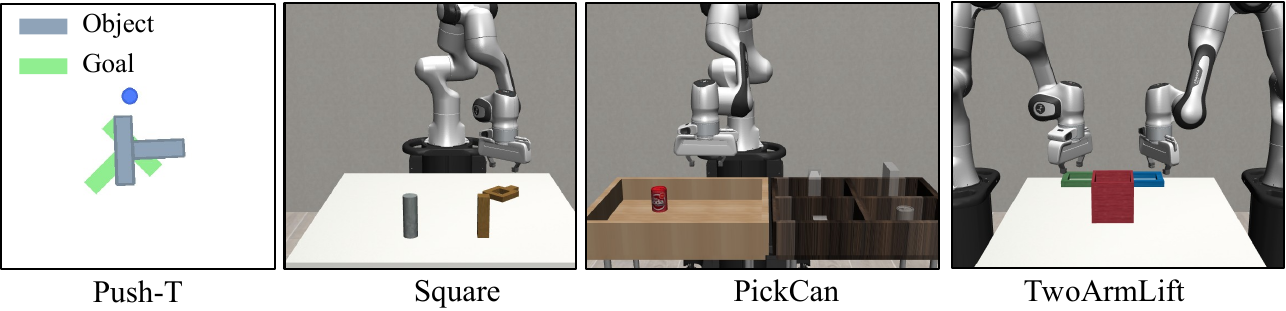}
	\caption{
Tasks in the simulation experiments.
 }
\label{fig:appendix_simulated_tasks}
\end{figure}

\subsubsection{Task descriptions}

The evaluated tasks are summarized below.
(i) Push-T: Originally introduced by \cite{2023_diffusionpolicy}, this task requires a robot to push a T-shaped object to a predefined target location using a circular end effector.
(ii) Square: The robot is required to align and place a square nut onto a fixed square peg.
(iii) PickCan: In this task, the robot must grasp a can and move it to a designated target bin.
(iv) TwoArmLift: Two robot manipulators collaboratively lift a single shared object.
Each observation consists of RGB images from cameras and the robot's end effector pose.
At the start of each episode, the initial position of the manipulated object is sampled uniformly at random.
For the Push-T task, instead of reporting the target area covered as in the original DP paper, we report the success rate of completing this task.

 \subsubsection{Action space}
 We consider two control modes: (1) For absolute position control, the single-step action is the pose of the robot end effector in the fixed world coordinate frame, along with a gripper command when applicable. 
(2) For velocity control, the single-step action is the delta change of the robot end effector pose and the gripper command.
To ensure stable absolute position control during online learning, we apply a norm-based saturation on the commanded end-effector displacement. If the norm of the displacement exceeds a threshold ($d=0.05$ for Robosutie and 15 for Push-T), it is rescaled to this maximum value while preserving its direction. 
This saturation is applied only to the end-effector position.
This saturation is disabled for offline learning with the Robomimic dataset to maintain consistency with the dataset’s action distribution.

\begin{figure*}[t!]
	\centering
	\includegraphics[width=1.0\textwidth]{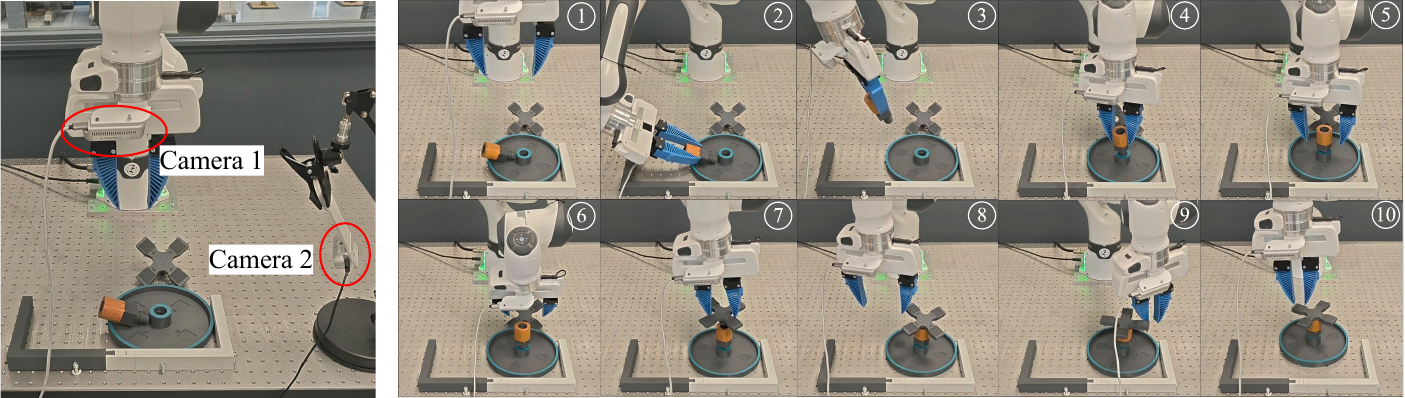}
	\caption{Left: Setup of the round-table experiment. Right: Example rollout of the SDP policy, trained with \textit{Demo30 + Corrections}, during evaluation.}
\label{fig:appendix_round_Table_rollout}
\end{figure*}

\subsubsection{Simulated Teachers and Feedback}
We employ a simulated teacher that monitors deviations between the robot action $\mathbf a^r$ and the optimal teacher's action $\mathbf a^*$, both defined in the same action space. Every 2 steps, if the difference exceeds a predefined threshold, the simulated teacher starts to give corrections for $2T$ consecutive steps and then gives the control back to the robot.
The deviation threshold is defined to 0.05 for absolute position control and 0.1 for velocity control.

For accurate feedback, the teacher action equals the optimal action: $\mathbf a^h = \mathbf a^* $.
For noisy feedback, we consider only absolute position control and model the teacher action as: 
\begin{align*}
\mathbf a^h = \mathbf a^* + \mathbf \omega,\quad
\mathbf \omega \sim \mathcal{N}
\bigl(
\mathbf 0,\;
\sigma^2 \mathbf I
\bigr),
\end{align*}
where
\begin{align*}
\sigma = \min\!\left(
\frac{1}{2}\lVert \mathbf a^* - \mathbf a^r \rVert,\; 0.04
\right).    
\end{align*}
The upper bound $\sigma \leq 0.04$ is imposed to prevent the injected noise from becoming excessively large, as excessively large noise would eliminate meaningful guidance from the teacher.

\subsubsection{Evaluation}
For both online and offline learning, we train 296 episodes for 3 trials with different seeds. For each trial, we save checkpoints every 5 episodes and evaluate the last 21 checkpoints, each on 10 episodes with distinct initial states and evaluation seeds.
We obtain the average of the success rates and report these results in Experiment section \ref{sec:exp:interactive_learning}, \ref{sec:exp:offline_learning_results}, and \ref{sec:exp:ablation}

\subsection{Details of Real-World Experiments}
\label{Appendix:details_of_real_world}

\subsubsection{Task description}
At each step, the observation $o_t$ includes RGB images from two cameras and the robot's end effector pose.
The policy conditions on the current observation together with a fixed number of past observations and outputs an action chunk of horizon $T$. 
During online interaction and evaluation, only the first $T_a$ single-step actions of the predicted action-chunk are executed. 

For the Insert-T task, the single-step action is the 2D position
over the table plane.
For the roundtable assembly task, the single-step action is the pose of the robot end effector in the fixed world coordinate frame and the gripper command. 
The pose includes position and orientation, where the position is normalized by a predefined workspace and the orientation is represented as the first two columns of a rotation matrix \cite{2023_diffusionpolicy}.
The gripper command is also normalized to a fixed range. 

The single-step actions are sent to the low-level controller at 10 Hz, and the robot policy outputs a new action-chunk once the first $T_a$ single-step actions of the previous action-chunk have been executed.
Each single-step action is de-normalized before being sent to the low-level controller.
For the Franka robot, we employ a Cartesian impedance controller operating at 1 kHz. For the KUKA robot, commanded Cartesian targets are first converted to joint-space goals using an inverse kinematics solver, and then tracked by a joint impedance controller running at 500 Hz.
An overview of the round-table assembly setup is shown in Fig. \ref{fig:appendix_round_Table_rollout}. 

\subsubsection{Data collection}
\label{appendix:data_collection}
We follow the pipeline illustrated in Fig. \ref{fig:framework}. 
Our dataset is collected from three sources: demonstrations, free-play, and online corrections.

We first collect \textbf{demonstration} data using a space mouse device.
This data serves as the initial pretraining dataset.
Meanwhile, we also collect a \textbf{free-play} dataset used only for training the observation autoencoder (Appendix \ref{appendix:obs_encoder}). 
During free-play, the operator freely teleoperates the robot around the task objects without a predefined goal. The operator deliberately explores both in-distribution and out-of-distribution states by interacting with and repositioning the objects.
Data is saved to the free-play dataset when the operator presses the `enter' key of the keyboard.

Using the demonstration and free-play datasets, we train the SDP agent for 12 hours on an Nvidia A40 GPU, with hyperparameters reported in Table \ref{tab:appendix_hyperparameters}. The trained agent is then deployed to perform the task, during which a human operator can provide online interventions via the space mouse to correct undesired behaviors.
According to the IIL loop in Algorithm \ref{alg:CDP_algorithm}, extracting contrastive action-chunk data requires intervention segments of length 
$T$. Given a control frequency of 10 Hz and an action horizon of 16, we instruct the operator to maintain each intervention for at least 1.6 s after initiating control.
In practice, there are a few cases where a teacher's action is missing from a $T$-window due to the operator mistakenly releasing the control too early. 
To handle these cases and avoid wasting intervention data, we use a Python script to visualize the trajectory data and highlight the missing actions. 
These missing teacher actions can be relabeled by setting $\mathbf{a}^h \leftarrow \mathbf{a}^r$ if the human operator observes that the corresponding robot action is close to the desired actions they would provide. The \textbf{correction} dataset is then extracted from the relabeled trajectories.

The resulting correction dataset is aggregated with the demonstration and free-play datasets to retrain the SDP agent. This deployment-and-correction cycle can be repeated iteratively. We perform one iteration for the Insert-T task and three iterations for the round-table assembly task.
Final dataset sizes, reported as numbers of observation–action pairs, are summarized in Table \ref{tab:appendix_dataset_size}.
\begin{table}[h]
\centering
\setlength{\tabcolsep}{8pt} % reduce column spacing
\renewcommand{\arraystretch}{1.1} % slightly tighter rows
\caption{Dataset size in real-world experiments}
\label{tab:appendix_dataset_size}
\begin{tabular}{lccc}
\Xhline{0.75pt}
& Initial  
& Free-play
& Correction 
\\
& Demonstration & dataset & \\
\hline
Insert-T & 10,925 & 2,011 & 10,110\\
Round-table     & 30,626 &2,222 & 29,194  \\
\Xhline{0.75pt}
\end{tabular}
\label{tab:gs_reflection_results}
\end{table}

\subsubsection{Evaluations}
Using the demonstration-only dataset, we train both SDP and DP for a fixed training budget of 12 hours on an Nvidia A40. From the final checkpoint of this stage, each method is further trained for an additional 12 hours using the combined demonstration and correction dataset. For each method and dataset setting, the final checkpoint after training is selected for evaluation.
% All evaluations are conducted by executing the learned policy from initial states sampled from the same distribution used during data collection for demonstrations and corrections.

\begin{figure}[h!]
	\centering
	\includegraphics[width=0.48\textwidth]{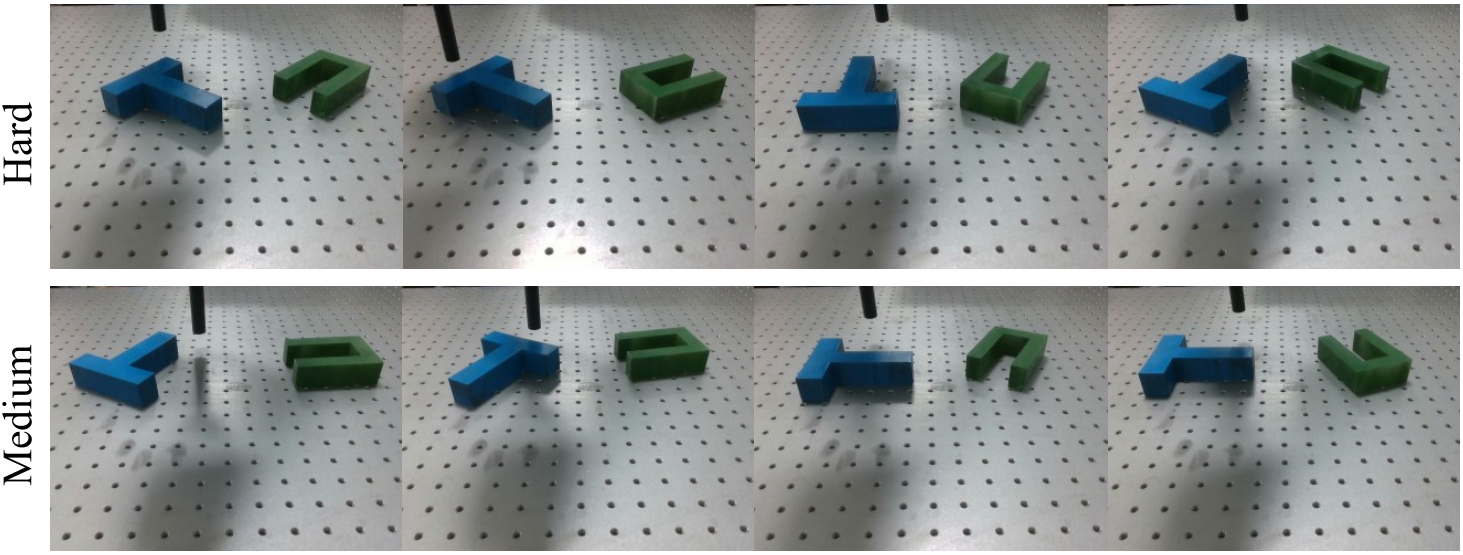}
	\caption{
Examples of the initial states during the evaluation of the Insert-T task, for both the hard and medium tasks.
 }
\label{fig:Appendix_InsertT_initial_states}
\end{figure}

For the Insert-T task, we evaluate the methods with the same set of 40 initial states for both hard and medium tasks. 
Examples of the initial states are shown in Fig. \ref{fig:Appendix_InsertT_initial_states}.

\begin{figure}[h!]
	\centering
	\includegraphics[width=0.48\textwidth]{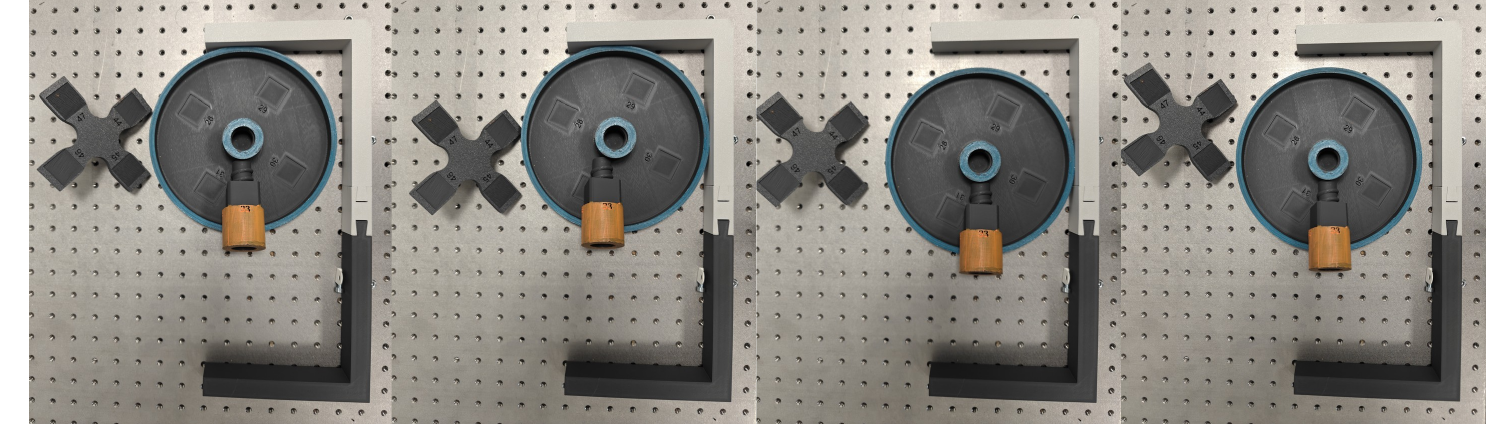}
	\caption{
Examples of the initial states during the evaluation of the round-table assembly task.
 }
\label{fig:Appendix_roundtable_initial_states}
\end{figure}

For the round-table assembly task, 20 initial states are evaluated for all methods. The orientation of the bottom part and the middle part are fixed in the global frame, while their positions are varied across initial states.
Representative initial states are shown in Fig. \ref{fig:Appendix_roundtable_initial_states}.
One example rollout of SDP agent during evaluation is shown in Fig. \ref{fig:appendix_round_Table_rollout}. 
% to do, say how we randamzie the states

%% file: tables/appendix_hyperparameter.tex
\begin{table*}
\footnotesize
\caption{Hyperparameters for Set-Supervised Diffusion Policy.
Ctrl: position or velocity control. 
$T_o$: observation horizon. 
$T$: action-chunk horizon. 
$T_a$: action execution horizon, of which $T_a$ steps of an action-chunk are executed.
$T_r$: queried action-chunk length during intervention.
$\text{ImgRes}$: environment observation resolution (Camera views x W x H).
$r$: the radius ratio of the single-step desired action set.
$N$: number of desired action-chunk samples.
$K$: total diffusion steps.
$K_A$: start denoising step for sampling desired action-chunks.
$K_{\text{inference}}$: total diffusion steps during inference.
 $N_{\textnormal{training}}$: number of training steps performed at the end of each episode.
% A ‘$\diagdown$’ symbol denotes that the algorithm did not converge. For calculating CT, $\diagdown$ entries are replaced with the maximum allowable timestep.
}
\label{tab:appendix_hyperparameters}
\begin{center}
\small
\begin{tabular}{l|llllllllllll}
\toprule
\textbf{Hyperparameters} & {Ctrl} & $T_o$ & $T$ & $T_a$ & $T_r$ & {ImgRes} & Radius ratio $r$ &  $N$& $K$& ${K_A}$  & $K_{\text{inference}} $   &$N_{\text{training}}$ \\
\midrule
\textbf{Online Accurate} && & &  & & & & & & &  \\
Push-T       & Pos  & 2  & 16  & 8 & 2   & 2x84x84   & 0.1      & 16       & 100 & 16  & 16    & 200           \\
Square       & Pos  & 2  & 16  & 8 & 2   & 2x84x84   & 0.1      & 16       & 100 & 16  & 16    & 200           \\
PickCan    & Pos  & 2  & 16  & 8 & 2   & 2x84x84   & 0.1      & 16       & 100 & 16  & 16    & 200           \\
TwoArmLift      & Pos  & 2  & 16  & 8 & 2   & 4x84x84   & 0.1      & 16       & 100 & 16  & 32    & 200           \\
\midrule
\textbf{Online Noisy} && & &  & & & & & & &  \\
Push-T       & Pos  & 2  & 16  & 8 & 2   & 2x84x84   & 0.6      & 16       & 100 & 16  & 16        & 200       \\  
Square     & Pos  & 2  & 16  & 8 & 2   & 2x84x84   & 0.6      & 16       & 100 & 16  & 16        & 200       \\  
PickCan     & Pos  & 2  & 16  & 8 & 2   & 2x84x84   & 0.6      & 16       & 100 & 16  & 16        & 200       \\  
TwoArmLift     & Pos  & 2  & 16  & 8 & 2   & 4x84x84   & 0.6      & 16       & 100 & 16  & 32        & 200       \\  
\midrule
\textbf{Offline Robomimic} && & &  & & & & & & &  \\
Position T= 1& Pos  & 2  & 1  & 1 & N/A  & 2x84x84   & 0.01         & 16       & 100 & 16  & 16        & 400       \\ 
Position T= 16& Pos  & 2  & 16  & 8 &N/A   & 2x84x84   & 0.01         & 16       & 100 & 16  & 16        & 200       \\ 
Velocity T= 1& Vel  & 2  & 1  & 1 & N/A  & 2x84x84   & 0.01         & 16       & 100 & 16  & 16        & 200       \\ 
Velocity T= 16& Vel  & 2  & 16  & 8 & N/A   & 2x84x84   & 0.01         & 16       & 100 & 16  & 16        & 200       \\ 
\midrule
\textbf{Offline SDP/DP-data} && & &  & & & & & & &  \\
Position T= 1& Pos  & 2  & 1  & 1 & N/A   & 2x84x84   & 0.1         & 16       & 100 & 16  & 16        & 200       \\ 
Position T= 16& Pos  & 2  & 16  & 8 & N/A   & 2x84x84   & 0.1         & 16       & 100 & 16  & 16        & 200       \\ 
Velocity T= 1& Vel  & 2  & 1  & 1 & N/A  & 2x84x84   & 0.1         & 16       & 100 & 16  & 16        & 200       \\ 
Velocity T= 16& Vel  & 2  & 16  & 8 & N/A   & 2x84x84   & 0.1         & 16       & 100 & 16  & 16        & 200       \\
\midrule
\textbf{Real-world Demonstration} && & &  & & & & & & &  \\
Insert-T & Pos  & 2  & 16  & 8 & N/A   & 2x320x240   & 0.01         & 16       & 100 & 16  & 16        & 200       \\ 
Round-table Assembly & Pos  & 2  & 16  & 8 & N/A   & 2x320x240   & 0.001         & 16       & 100 & 16  & 16        & 200       \\ 
\midrule
\textbf{Real-world Correction Dataset} && & &  & & & & & & &  \\
Insert-T & Pos  & 2  & 16  & 8 & 2   & 2x320x240   & 0.1         & 16       & 100 & 16  & 16        & 200       \\ 
Round-table Assembly & Pos  & 2  & 16  & 8 & 2   & 2x320x240   & 0.01         & 16       & 100 & 16  & 16        & 200       \\ 
\bottomrule
\end{tabular}
\end{center}
\end{table*}